\documentclass[11pt,letterpaper]{Sengoku}
\usepackage{amsmath,amsfonts,bm}

\def\eqref#1{equation~\ref{#1}}

\def\1{\bm{1}}

\DeclareMathAlphabet{\mathsfit}{\encodingdefault}{\sfdefault}{m}{sl}
\SetMathAlphabet{\mathsfit}{bold}{\encodingdefault}{\sfdefault}{bx}{n}

\DeclareMathOperator*{\argmax}{arg\,max}

\usepackage{booktabs}
\usepackage{longtable,array}
\usepackage{wrapfig}
\usepackage{capt-of}
\usepackage{listings}
\usepackage{url}
\tcbset{promptpanel/.style={enhanced,breakable,colback=white,colframe=black!50,colbacktitle=black!50,coltitle=white,boxrule=0.6pt,arc=4pt,title filled,fonttitle=\rmfamily\bfseries\large,left=8pt,right=8pt,top=7pt,bottom=7pt,toptitle=4pt,bottomtitle=4pt,before skip=10pt,after skip=10pt}}
\newtcolorbox{agentprompt}[1]{promptpanel,title={#1},title after break={#1\enspace\textnormal{(continued)}},fontupper=\fontfamily{cmtt}\fontsize{9}{10.5}\selectfont\raggedright,before upper={\setlength{\parindent}{0pt}\setlength{\parskip}{4pt}},lines before break=5}
\newtcblisting{agentpromptlisting}[1]{promptpanel,title={#1},listing only,listing options={basicstyle=\fontfamily{cmtt}\selectfont\small,numbers=none,breaklines=true,breakatwhitespace=true,columns=fullflexible,keepspaces=true,showstringspaces=false,aboveskip=0pt,belowskip=0pt}}

\usepackage{marvosym}

\ExplSyntaxOn
\prop_new:N \g_waa_author_mark_prop
\NewDocumentCommand \authormark { m m }
  { \prop_gput:Nnn \g_waa_author_mark_prop {#1} {#2} }
\cs_set_protected:Npn \sengoku_author_item:nn #1#2
  {
    \hbox_set:Nn \l_sengoku_author_box
      {
        \strut
        \sengoku_author_badges:n {#2}
        {\bfseries #1}
        \tl_if_blank:nF {#2}
          {
            \nobreak\textsuperscript
              {
                \sengoku_affiliation_numbers:n {#2}
                \prop_item:Nn \g_waa_author_mark_prop {#1}
              }
          }
      }
    \mode_leave_vertical:
    \dim_compare:nNnTF {\box_wd:N \l_sengoku_author_box} > {\linewidth}
      {\hbox_unpack_drop:N \l_sengoku_author_box}
      {\box_use_drop:N \l_sengoku_author_box}
  }
\ExplSyntaxOff

\graphicspath{{Figures/}}
\title{\gradienttitle{World}{Action}{Agent:}{Harnessing VLMs for Robot Manipulation via World Action Rehearsal}}
\date{\today}

\shorttitle{World Action Agent}
\runningauthor{Yehang Zhang et al.}

\affiliationlayout{inline}
\affiliationgap{1.0em}
\affiliation{hkustgz}{HKUST(GZ)}
\affiliation{cuhk}{CUHK}
\affiliation{knowin}{Knowin AI}

\author[hkustgz,knowin]{Yehang Zhang}
\author[hkustgz,knowin]{Haojian Huang}
\author[knowin]{Yifan Chang}
\author[hkustgz,knowin]{Jianchong Su}
\author[cuhk,knowin]{Bohan Zhou}
\author[hkustgz,knowin]{Yingjie Xu}
\author[hkustgz,knowin]{Wosong Chen}
\author[hkustgz,knowin]{Tianhao Zhou}
\author[knowin]{Chenxu Wang}
\author[knowin]{Tianyi Zhang}
\author[knowin]{Yangkai Wei}
\author[cuhk,knowin]{Wenqian Li}
\author[knowin]{Shiyuan Deng}
\author[knowin]{Yinchuan Li}
\author[hkustgz]{Ying-Cong Chen}
\author[cuhk,knowin]{Zexi Li}

\authormark{Yehang Zhang}{\,\ensuremath{*}}
\authormark{Haojian Huang}{\,\ensuremath{*}}
\authormark{Zexi Li}{\,\ensuremath{\dagger}}
\authormark{Ying-Cong Chen}{\,\Letter}
\makeatletter
\gdef\Sengoku@titlenotes{%
  \textsuperscript{\ensuremath{*}}\,Equal contribution.\quad
  \textsuperscript{\ensuremath{\dagger}}\,Project leader.\quad
  \textsuperscript{\Letter}\,Corresponding author.\par}
\makeatother

\hypersetup{
  pdfsubject={Research article}
}

\begin{document}

\begin{abstract}
General-purpose vision-language models (VLMs) bring broad knowledge and spatial reasoning to robot manipulation, yet existing systems either use them indirectly, to predict constraints or write programs, or give them a view of the scene rather than a world in which to act.
We present \textbf{World Action Agent (WAA)}, a multi-agent harness through which VLMs pilot robots with basic tools, making every decision within a \textbf{visual action workspace}.
The workspace has three properties.
\textbf{Contact views}, selected automatically from the scene geometry, present the scene around the current interaction.
\textbf{Action rehearsal} turns each action into an editable proposal that the agent, alone or through an Imagination Agent, previews and revises against planning feedback before execution.
\textbf{In-view correction} closes the loop between observation, rehearsal, and low-level execution, letting the agent remove residual offsets in the view where it observes them.
Through the same workspace, WAA acquires embodied procedural knowledge in two ways: it evolves multimodal skills from expert videos and human teaching under evidence-based review and consults them through a Skill Agent, and its interaction traces train smaller VLMs to pilot the same harness.
On LIBERO-Pro, WAA with skills evolved only from LIBERO-90 reaches a state-of-the-art \textbf{75.6\%} average success, outperforming end-to-end VLAs, code-as-policy agents, and a visual-harness baseline with the same backbone; the same skills remain effective on robosuite without further learning.
Fine-tuning Qwen3.5-9B on harness traces raises its out-of-domain success from 1.7\% to 43.3\%.
\end{abstract}

\maketitle

\section{Introduction}
\label{sec:introduction}

Learning manipulation policies that generalize to new objects, layouts, and tasks is a central goal in robotics.
Vision-language-action (VLA) models learn such policies from visual observations, language instructions, and robot actions \citep{kim2024openvla,intelligence2025pi_}, but fine-tuning vision-language models (VLMs) for action prediction may weaken their general understanding and reasoning \citep{hancock2026actions}.
World action models (WAMs) couple environment dynamics with action learning \citep{zhu2025unifiedworldmodelscoupling,ye2026world}, yet grounding learned visual dynamics in executable control still requires robot data and action alignment \citep{wam-survey-nus}.
A complementary route preserves the generality of VLMs and involves them in manipulation decisions.
HAMSTER and ReKep have the VLM predict intermediate paths or relational constraints that a downstream policy or optimizer turns into motions \citep{li2025hamster,huang2024rekep}; Code as Policies (CaP) and CaP-X have a language model compose perception and control APIs into programs and revise them from execution feedback \citep{liang2023code,fu2026capx}.
In both cases, the VLM either supplies constraints and spatial guidance or indirectly writes a program, rather than controlling the robot directly through its own spatial understanding and basic primitives.

Show-Harness and VIA take a step in this direction: the VLM selects and revises actions directly through a visual robot interface, turning manipulation into an interactive visual reasoning problem \citep{chen2026show,hu2026via}.
Yet such interfaces display the scene to the VLM and let it choose actions; they do not give it a world in which to act.
Three things are missing.
First, observation is not centered on the interaction, although success depends on local relations among the gripper, the object, and the target.
Second, actions cannot be tried before they are taken: each takes effect as soon as it is issued, before the VLM can see its consequences.
Third, perception and action live in different spaces: an offset observed in an image must be rewritten as coordinates before it can be corrected.
This motivates our question: \emph{how can a general-purpose VLM use basic action primitives to make and revise decisions throughout execution, serving as a robot pilot?}

\begin{figure}[t!]
    \centering
    \includegraphics[width=\textwidth]{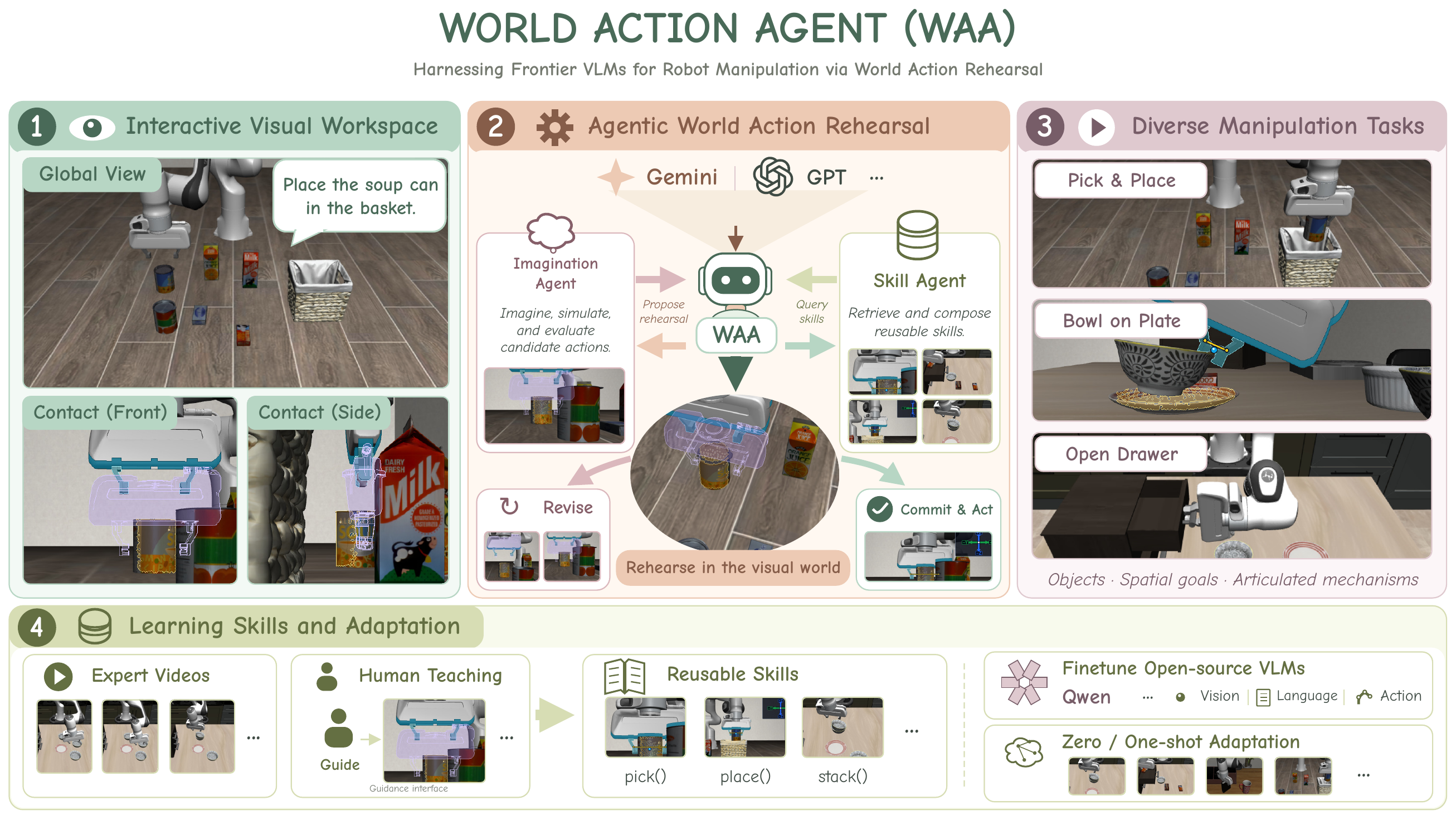}
    \caption{\textbf{World Action Agent (WAA).} WAA combines visual action rehearsal with reusable skills, enabling VLMs to perform diverse manipulation tasks and adapt through expert demonstrations and human guidance.}
    \label{fig:teaser}
\end{figure}

Our answer is to change what the VLM sees and how its decisions take effect, rather than the VLM itself.
We introduce \textbf{World Action Agent (WAA)} (Figure~\ref{fig:teaser}), a multi-agent embodied harness in which agents observe the scene, construct actions, and drive the robot through basic tools alone, build and retrieve multimodal skills, and make every decision within a single \textbf{visual action workspace}.
On this basis, WAA has three properties.
First, it presents the scene around the interaction: Contact views are selected automatically for the current interaction, letting the VLM bring its spatial understanding to bear.
Second, actions can be rehearsed and revised before execution: each action is an editable proposal with planning feedback.
Third, observation, rehearsal, and low-level execution form a closed loop within the same set of calibrated views.

Piloting a robot also requires embodied procedural knowledge that VLM pretraining rarely captures.
WAA acquires this knowledge through the same workspace in two ways: it evolves multimodal skills from expert videos and human teaching, and it uses interaction traces to train a smaller VLM to pilot the same harness.

On LIBERO-Pro \citep{zhou2026liberoprorobustfairevaluation}, WAA with skills evolved only from LIBERO-90 reaches 75.6\% average success, above ASPIRE (72.0\%), and the same skills remain effective on robosuite \citep{zhu2025robosuitemodularsimulationframework} without further learning.
Fine-tuning Qwen3.5-9B \citep{qwen35blog} on harness traces raises its out-of-domain success from 1.7\% to 43.3\%.
Our main contributions are:
\begin{itemize}
    \item \textbf{World Action Agent.}
    We introduce WAA, a multi-agent harness through which general-purpose VLMs pilot robots with basic tools, observing, rehearsing, and correcting every action within a single visual action workspace.
    \item \textbf{Learning through the harness.}
    We show that the same workspace supports both non-parametric skill evolution from expert videos and human teaching and the distillation of interaction traces into smaller VLM pilots.
    \item \textbf{State-of-the-art results.}
    WAA achieves a state-of-the-art 75.6\% average success on LIBERO-Pro and transfers to robosuite without further learning. Further analyses show that a 9B VLM can learn to pilot the same harness.
\end{itemize}

\section{Related Work}
\label{sec:related_work}

\noindent\textbf{Foundation models for robot manipulation.}
Foundation models support robot manipulation from action prediction to high-level decision-making.
Vision-language-action (VLA) models, including RT-2, OpenVLA, and $\pi_{0.5}$, generate robot actions from vision and language through discrete tokens or continuous action modules \citep{brohan2023rt,kim2024openvla,intelligence2025pi_}.
However, fine-tuning VLMs for action prediction may weaken their general reasoning and multimodal understanding, limiting generalization \citep{hancock2026actions}.
World action models (WAMs) couple future-state prediction with robot action learning.
Recent approaches jointly model video and actions or adapt pretrained video models for control \citep{li2025unifiedvideoactionmodel,zhu2025unifiedworldmodelscoupling,bi2026motus,kim2026cosmospolicyfinetuningvideo,li2026causalworldmodelingrobot,ye2026world}.
Others reduce inference cost by omitting future-video generation or predicting compact latent futures \citep{yuan2026fastwamworldactionmodels,lin2026jepawamlearningvisionlanguageactionpolicies}.
However, executable control typically requires robot data for action alignment, and cross-embodiment transfer remains sensitive to morphology and action-space differences \citep{wam-survey-nus}.
Complementary work uses VLM-generated spatial representations to guide robot control.
HAMSTER uses a fine-tuned VLM to predict coarse 2D end-effector paths that guide a 3D-aware low-level policy \citep{li2025hamster}.
ReKep generates relational keypoint constraints for hierarchical optimization \citep{huang2024rekep}, while OmniManip grounds VLM reasoning in object-centric interaction primitives that specify 3D spatial constraints \citep{11095109}.
FSD trains a VLM to use spatial reasoning to generate affordance regions, points, and visual traces for motion planning \citep{yuan2026from}.
These approaches use VLM knowledge and spatial reasoning to guide actions, supporting VLMs as core decision makers for manipulation.

\noindent\textbf{Agentic robot control and visual interfaces.}
Code as Policies (CaP) and subsequent code-generation systems connect language-model reasoning to robot execution through programs over perception and control APIs \citep{liang2023code,10161317,chen2024roboscriptcodegenerationfreeform,3692070.3693552}.
Recent systems incorporate interactive execution and experience reuse.
CaP-X evaluates and improves coding agents through multi-turn feedback and skill synthesis \citep{fu2026capx}.
RATs acquires reusable code skills through self-directed play \citep{rats2026playful}, while ASPIRE combines execution diagnosis, program repair, and evolutionary search to build transferable skill libraries \citep{lu2026aspireagenticskillsdiscovery}.
GaP structures robot programs as directed execution graphs and refines their structure and parameters through simulation rehearsal \citep{chen2026gapgraphaspolicymultiagentselflearning}.
Agent as Policy keeps a coding agent in the control loop to revise programs and motions from physical feedback \citep{jia2026agentaspolicy}.
Their effectiveness depends on the APIs' perceptual grounding and control abstractions: CaP-X reports declining performance as hand-designed abstractions are removed \citep{fu2026capx}.
Hierarchical and agentic systems also couple high-level reasoning with VLA execution.
Hi Robot, AtomBridge, and Harness VLA support this coupling through subtask instructions, inter-skill transitions, and memory-guided orchestration, respectively \citep{shi2025hi,pang2026atombridgeagenticvlainference,zhang2026harnessvlasteeringfrozen}.
These systems improve task composition, but each VLA call remains bounded by the learned policy's capabilities.
Visual interfaces such as VIA and Show-Harness instead expose spatial targets or fine-grained semantic actions directly to the VLM \citep{hu2026via,chen2026show}.
WAA gives the VLM direct control over robot primitives, with a visual action workspace for inspecting and revising actions before execution.
Numerical planners and controllers realize these decisions without a VLA executor or generated policy code.

\section{World Action Agent}
\label{sec:methods}

\begin{figure}[!ht]
    \centering
    \includegraphics[width=\textwidth]{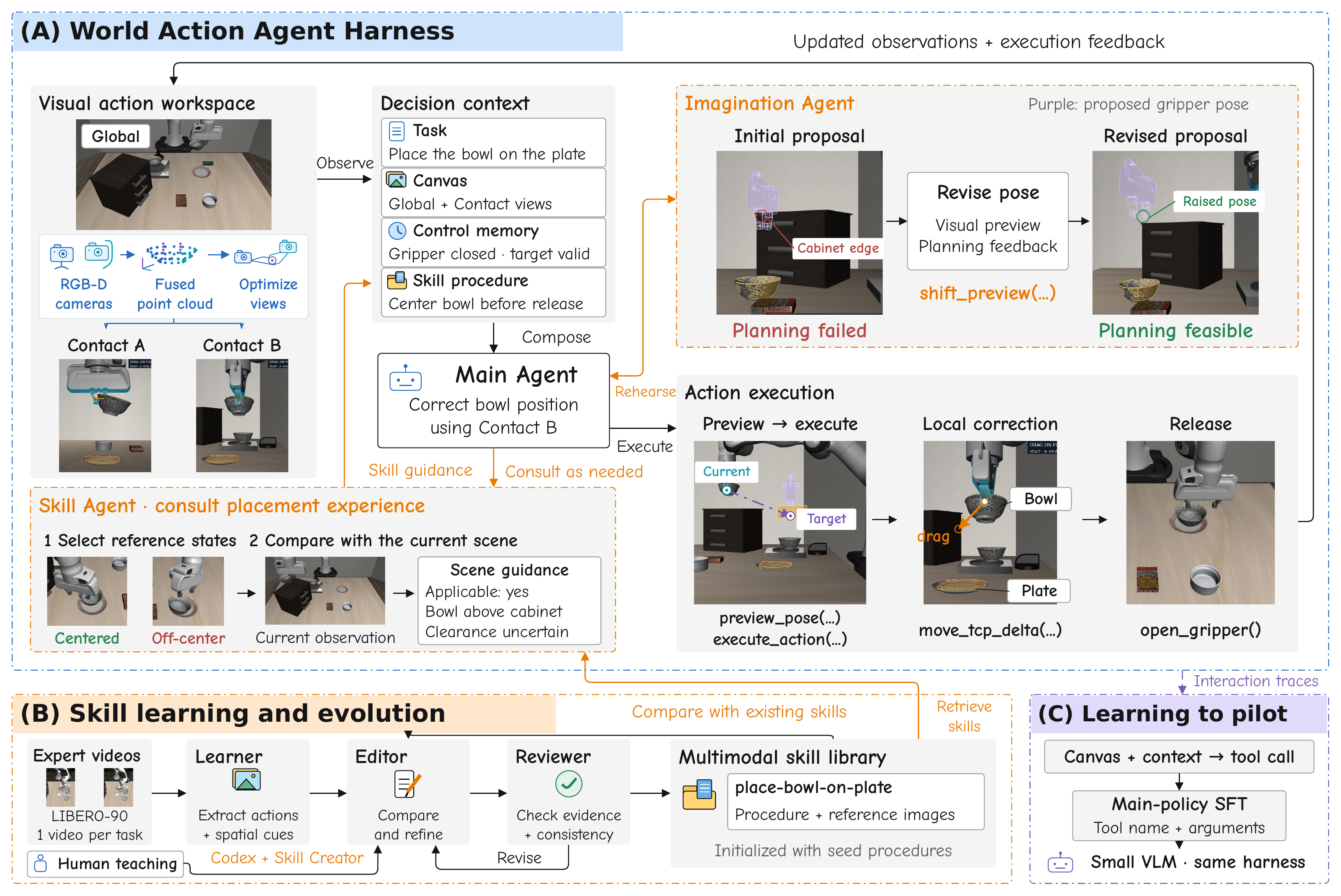}
    \caption{\textbf{Overview of the World Action Agent (WAA) harness.}
    \textbf{(A)} The main agent observes global and Contact views in the visual action workspace, rehearses action proposals with an Imagination Agent, consults skills through a Skill Agent, and executes planned motions and local corrections.
    \textbf{(B)} Expert videos and human teaching are distilled into a multimodal skill library through extraction, editing, and review.
    \textbf{(C)} Interaction traces train a smaller VLM to operate the same harness.
    Insets are from two recorded bowl-manipulation tasks; dashed arrows indicate target guides.}
    \label{fig:waa-method}
\end{figure}

\subsection{Problem Formulation}
\label{sec:problem-formulation}

Given a task instruction $g$, the agent completes a manipulation task through a sequence of tool calls.
At step $t$, the harness presents a multi-view Canvas $C_t$ and a control context $m_t$ that records valid spatial references, the pending action proposal $\hat{a}_t$, and recent execution feedback.
The policy then selects a tool call
\begin{equation}
    u_t \sim \pi_\theta\!\left(\cdot \mid g, C_t, m_t\right),
    \label{eq:agent-decision}
\end{equation}
where $u_t$ specifies a tool and its arguments.
Tools fall into three classes by their effect: \textbf{query} tools acquire spatial information or skill guidance, \textbf{proposal} tools construct and revise $\hat{a}_t$ and return a visual preview with planning feedback, and \textbf{execution} tools move the robot.
Because query and proposal tools leave the physical world unchanged, the agent can inspect and revise an action repeatedly before committing to it.
The episode ends when the task succeeds or the interaction budget is exhausted.
Within this formulation, the VLM decides what to manipulate, how to revise a proposal, and when to execute it, while the harness turns these decisions into geometrically accurate and kinematically feasible robot motion.

\subsection{Designing the World Action Agent Harness}
\label{sec:harness-design}

\noindent\textbf{Design principle.}
General-purpose VLMs excel at semantic understanding and qualitative spatial judgment, but they struggle to produce precise metric poses and cannot anticipate whether an action is physically executable.
When a VLM drives a robot directly, three difficulties arise: local relations among the gripper, object, and target are hard to discern from a global view; physical actions are irreversible and cannot be verified in advance; and planned motions leave fine alignment errors.
WAA addresses them with three corresponding designs that together form a \textbf{visual action workspace} (Figure~\ref{fig:waa-method}A): an interaction-centered Canvas lets the agent see local relations, action rehearsal lets it test actions before execution, and in-view correction lets it remove residual errors from new observations.
Because all three share one calibrated Canvas, the agent observes, rehearses, and corrects the same spatial targets.

\noindent\textbf{Interaction-centered Canvas.}
Manipulation outcomes often hinge on millimeter- to centimeter-scale relations among the gripper, object, and target, which a fixed global camera frequently fails to reveal because of distance, occlusion, or degenerate viewing angles.
WAA therefore selects viewpoints actively according to the current interaction.
Given the scene point cloud $\mathcal{P}_t$, the robot geometry $\mathcal{R}_t$, and the highlighted interaction region $\mathcal{H}_t$, the harness solves for the camera parameters of the \textbf{Contact views}
\begin{equation}
    \mathbf{c}_t=\argmax_{\mathbf{c}\in\Omega}\;
    S_{\mathrm{vis}}(\mathbf{c};\mathcal{H}_t,\mathcal{R}_t\mid\mathcal{P}_t)
    +\lambda_1 S_{\mathrm{frame}}(\mathbf{c};\mathcal{H}_t)
    -\lambda_2 E_{\mathrm{red}}(\mathbf{c})
    -\lambda_3 E_{\mathrm{stab}}(\mathbf{c},\mathbf{c}_{t-1}),
    \label{eq:contact-view-selection}
\end{equation}
where $S_{\mathrm{vis}}$ measures the visibility of the interaction region and gripper under scene occlusion, $S_{\mathrm{frame}}$ encourages compact framing, $E_{\mathrm{red}}$ discourages views that convey the same direction, and $E_{\mathrm{stab}}$ suppresses viewpoint jumps so that spatial relations remain comparable across steps.
The feasible set $\Omega$ constrains the horizontal projections of the two Contact views to be orthogonal, so that an alignment error can be read along two independent directions.
The objective only requires a scene point cloud in the robot base frame, so WAA is agnostic to how $\mathcal{P}_t$ is obtained: it can be rendered in simulation, fused from calibrated RGB-D cameras together with the forward-kinematic hand geometry, or reconstructed from RGB views with a feed-forward model such as VGGT~\citep{11094896}.
Together with the global view, the optimized Contact views form the Canvas: the global view provides task context, the Contact views expose fine interaction relations, and every view retains its projection calibration so that image-space annotations map directly to 3D.

\begin{figure}[!t]
    \centering
    \includegraphics[width=\textwidth]{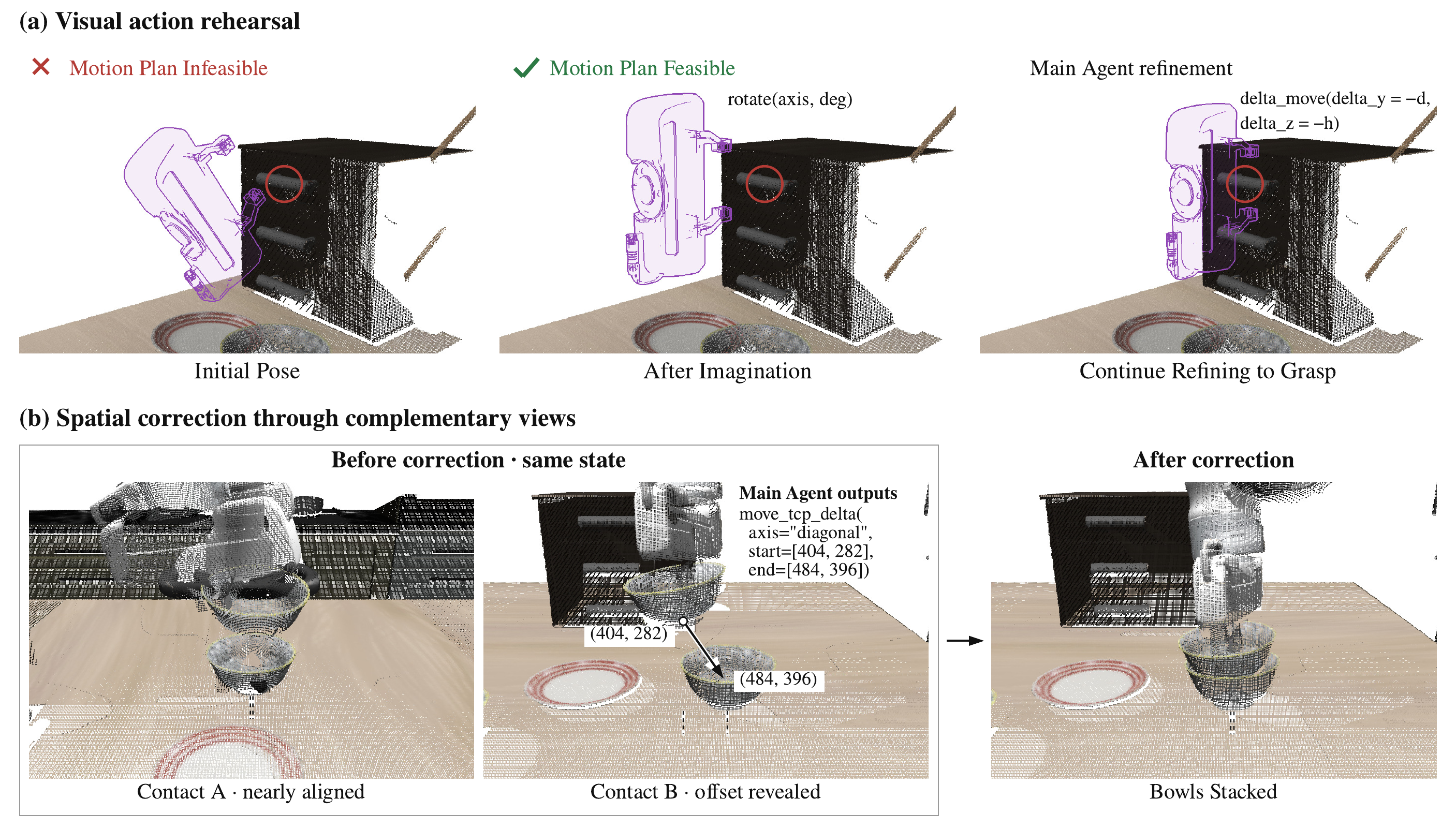}
    \caption{\textbf{Action rehearsal and in-view correction.}
    (a) The Imagination Agent rotates an infeasible grasp preview (purple) until planning succeeds; the main agent then shifts it toward the handle (red circle).
    (b) Contact A appears aligned, but Contact B reveals an offset; the main agent drags the bowl toward its target in Contact B, and the harness converts the drag into end-effector motion.
    These interface examples are rendered from RGB-D point clouds.}
    \label{fig:waa-rgbd-actions}
\end{figure}

\noindent\textbf{Action rehearsal.}
For a VLM, the hardest part is often not deciding where to go but judging whether a specific pose is appropriate: whether the approach collides with nearby objects, whether the robot can reach it, and whether the resulting grasp supports the next step.
WAA therefore represents an action as an \textbf{editable proposal} rather than a one-shot output.
The agent localizes targets on the Canvas and specifies a target pose from them; the harness solves inverse kinematics, plans the motion with cuRobo~\citep{10160765}, overlays the target configuration as a translucent robot in every view, and reports its feasibility (Figure~\ref{fig:waa-rgbd-actions}a).
The agent can thus examine approach direction, prospective contact, and surrounding clearance before execution and revise the pose accordingly.
For finer adjustment, the main agent delegates its spatial intent to an \textbf{Imagination Agent}, which runs in a separate context and iteratively edits and replans the proposal while the physical scene remains unchanged, returning only the revised proposal.
Only a proposal with an executable plan can become physical motion.
Rehearsal thus couples the VLM's judgment of task geometry with the planner's feasibility checks, exposing errors before they occur in the physical world.

\noindent\textbf{Closed-loop execution and in-view correction.}
Target-level planned motions suit large movements such as approaching and transporting, but depth noise, calibration error, and contact disturbances leave residual errors that often decide whether placement succeeds.
WAA lets the agent express corrections directly \textbf{in the view where the error is observed}.
The agent drags from a reference point to its desired location in a Contact view, and the harness uses the view's calibration to convert this image-space intent into a bounded end-effector displacement (Figure~\ref{fig:waa-rgbd-actions}b).
The reference point can be the gripper or a visible point on the held object, so the agent can state where the object should go without first converting that relation into an absolute gripper pose.
The two orthogonal Contact views support corrections along complementary directions.
After each correction, a fresh observation lets the agent decide whether to adjust further, release the object, or proceed to the next target-level action.

\subsection{Learning and Evolution through the Harness}
\label{sec:skill-learning}

The visual action workspace in Section~\ref{sec:harness-design} lets a VLM observe and act on spatial relations, but it does not tell the agent how to use these capabilities to complete a task.
This requires embodied procedural knowledge that is rarely captured by VLM pretraining, such as which contact to establish first, which relation to verify before release, and how to recover from a failed grasp.
WAA acquires this knowledge through the same interface along two complementary paths.
Non-parametrically, the agent evolves a library of multimodal skills from expert demonstrations and human teaching and consults it during execution with its parameters fixed.
Parametrically, interaction traces recorded in the harness train a smaller VLM to pilot the same harness.

\noindent\textbf{Multimodal manipulation skills.}
Textual procedures alone struggle to convey judgments such as when an alignment is sufficient, and visual references can supply the evidence needed for state recognition and verification~\citep{zhang2026mmskillsmultimodalskillsgeneral,jiang2026visualskill}.
Each WAA skill therefore combines applicability conditions, a procedure, key states with reference images, and observable outcome checks (Appendix~\ref{app:skill-examples}).
The procedure specifies the relations that the gripper, object, and target should satisfy, the required contact and clearance, and how to recover from failure, rather than fixed coordinates or displacements.
Objects, target positions, and motion magnitudes are always resolved in the current scene, and numerical values appear only as tentative ranges to be adjusted from observed feedback.
In contrast to skill libraries that register validated per-object parameters such as grasp height offsets and yaw angles~\citep{lu2026aspireagenticskillsdiscovery}, this \textbf{relational guidance} lets a skill transfer across object instances, layouts, and viewpoints.

During execution, the main agent selects a skill by its name and description and delegates its interpretation to a \textbf{Skill Agent}.
Running in a separate context, the Skill Agent selects the relevant procedural states and reference images, compares them with the current Canvas, and reports applicability, scene differences, and relations that cannot yet be confirmed (Appendix~\ref{app:skill-system-prompt}).
The main agent receives the full textual procedure, whereas reference images remain in the Skill Agent's context, which prevents historical images from being mistaken for the current scene and keeps the main context compact.
Scene-specific advice expires when its associated observation or control state changes, so the agent does not act on outdated guidance.

\noindent\textbf{Skill evolution from demonstrations and teaching.}
The library starts from text-only seed skills that Codex (GPT-5.5) writes from the harness API, providing semantic-level guidance without hard-coded parameters or fixed action sequences.
We then enrich it with one expert trajectory per LIBERO-90 source task, whose scenes differ from those used for evaluation.
Learning proceeds through three agent roles, each driven by GPT-5.5 (Figure~\ref{fig:waa-method}B).
The \textbf{Learner} segments each video at observed state changes, interprets contact, object motion, and support changes in terms of harness actions, and extracts knowledge candidates that cite their source frames.
It does not read the existing skill text, so new evidence is not steered by prior conclusions.
The \textbf{Editor} compares each candidate with existing procedures and reference images, then adds, revises, retires, or defers skills with source frames as visual evidence.
The \textbf{Reviewer} independently checks the revised skills against the source evidence and returns concrete issues to the Editor, so that only supported changes that preserve existing applicability conditions enter the library.
Requiring evidence and independent review for every edit keeps the library from degrading as learning proceeds.

Demonstrations, however, show successful behavior and rarely reveal how to recover from mistakes.
WAA therefore provides a Canvas-GUI: humans see the same multi-view Canvas as the agent and operate the robot with the same tools, such as pointing in a view, dragging in a Contact view, and previewing and executing proposals.
Humans thus pilot the robot much as a GUI agent would, and the harness records these explicit actions in the same format as agent interactions.
The Learner contrasts the agent's attempt before failure with the human correction from the same state, identifies which relation, such as contact location, alignment direction, or release timing, accounts for the difference in outcome, and distills it into recovery guidance, which then passes through the same Editor--Reviewer loop as video-derived candidates.

\noindent\textbf{Learning to pilot.}
Large proprietary VLMs pilot the harness effectively, but at relatively high cost and latency.
We therefore distill interaction traces into a smaller VLM that operates the same harness (Figure~\ref{fig:waa-method}C).
Training examples come from successful episodes, retaining only clean executions without redundant operations.
We further include separately reviewed recovery segments, such as regrasping after an empty grasp, so that the model also learns to revise decisions from feedback.
Human corrections recorded through the Canvas-GUI share the format of agent traces and can likewise serve as supervision for recovery behavior.
Each example pairs the pre-action Canvas, task instruction, control context, and available skill guidance with the executed tool call, without the teacher's reasoning or any privileged simulator state.
The model is trained by maximum likelihood on the policy in Eq.~(\ref{eq:agent-decision}):
\begin{equation}
    \max_\theta \sum_{(g,C_t,m_t,u_t)\in\mathcal{D}} \log \pi_\theta\!\left(u_t \mid g, C_t, m_t\right).
    \label{eq:pilot-sft}
\end{equation}
Fine-tuning updates only the main policy, while the visual action workspace and sub-agents remain unchanged, so the model learns to use the interface rather than to replace it.

\section{Experiments}
\label{sec:experiments}

\subsection{Experimental Setup}
\label{sec:experiment-details}

\noindent\textbf{Benchmarks.}
We evaluate on LIBERO-Pro \citep{zhou2026liberoprorobustfairevaluation} and robosuite \citep{zhu2025robosuitemodularsimulationframework}.
Following CaP-X and Playful \citep{fu2026capx,rats2026playful}, we use six LIBERO-Pro splits, Object, Goal, and Spatial, each under initial-position swaps (Pos.) and task perturbations (Task), with ten tasks per split.
Each WAA setting is evaluated for 60 episodes per split, six per task.
In robosuite, we use cube lifting, stacking, and restacking.

\noindent\textbf{Evaluation protocol.}
We evaluate WAA with Gemini~3.7 Flash as the backbone in three settings: without skills (zero-shot), with the text-only seed skills, and with the evolved skill library.
The skill library is learned only from LIBERO-90 and frozen before evaluation: no LIBERO-Pro or robosuite rollout, failure log, or label updates skills, prompts, or model parameters.
Each episode starts in a fresh process and is limited to 50 main-agent turns, 50 physical operations, and one hour, with at most six turns per Imagination Agent call.
WAA accepts scene point clouds from multiple sources, including the simulator, fused RGB-D views, and VGGT reconstruction \citep{11094896}; Table~\ref{tab:libero-pro-main} reports all three.

\noindent\textbf{Baselines.}
We compare with three families of methods.
\emph{End-to-end VLAs}: OpenVLA \citep{kim2024openvla} and $\pi_{0.5}$ \citep{intelligence2025pi_}.
\emph{Code-as-policy agents}: CaP-Agent0 \citep{fu2026capx}, RATs with play-learned skills \citep{rats2026playful}, and ASPIRE \citep{lu2026aspireagenticskillsdiscovery}.
\emph{Visual agent harness}: Show-Harness \citep{chen2026show} (Gemini~3.7 Flash), which we rerun for 30 episodes per split.
Other results are taken from the original papers.

\noindent\textbf{Pilot fine-tuning.}
Unlike the skill library, the smaller pilot is trained on LIBERO-Pro: 1{,}774 main-agent tool calls from 112 successful Gemini~3.7 Flash episodes with evolved skills across all six splits.
We fine-tune Qwen3.5-9B \citep{qwen35blog} with LoRA (rank 8, 10 epochs, learning rate $5\times10^{-5}$; Appendix~\ref{app:sft}); at evaluation, it replaces only the main agent, while sub-agents and grounding still use Gemini~3.7 Flash.

\subsection{Main Results}
\label{sec:main-results}

\begin{table}[!t]
    \setlength{\abovecaptionskip}{0pt}
    \setlength{\belowcaptionskip}{4pt}
    \centering
    \caption{Success rates (\%) on LIBERO-Pro. Bold marks the best result in each column.}
    \label{tab:libero-pro-main}
    \small
    \setlength{\tabcolsep}{5pt}
    \renewcommand{\arraystretch}{1.05}
    \begin{tabular}{lrrrrrrr}
        \toprule
        & \multicolumn{2}{c}{Object} & \multicolumn{2}{c}{Goal} & \multicolumn{2}{c}{Spatial} & \\
        \cmidrule(lr){2-3}\cmidrule(lr){4-5}\cmidrule(lr){6-7}
        Method & Pos. & Task & Pos. & Task & Pos. & Task & Avg. \\
        \midrule
        OpenVLA & 0.0 & 0.0 & 0.0 & 0.0 & 0.0 & 0.0 & 0.0 \\
        $\pi_{0.5}$ & 17.0 & 1.0 & 38.0 & 0.0 & 20.0 & 1.0 & 12.8 \\
        CaP-Agent0 (CaP-X) & 22.0 & 18.0 & 26.0 & 17.0 & 12.0 & 14.0 & 18.2 \\
        CaP-Agent0 (Playful) & 27.0 & 31.0 & 29.0 & 16.0 & 13.0 & 23.0 & 23.2 \\
        RATs (Playful) & 61.0 & 63.0 & 43.0 & 36.0 & 29.0 & 31.0 & 43.8 \\
        ASPIRE & \textbf{98.0} & \textbf{95.0} & \textbf{81.0} & 45.0 & 51.0 & 60.0 & 72.0 \\
        Show-Harness & 13.3 & 16.7 & 0.0 & 10.0 & 0.0 & 0.0 & 6.7 \\
        \midrule
        \multicolumn{8}{l}{\textit{Scene point cloud}} \\
        WAA (zero-shot) & 53.3 & 46.7 & 30.0 & 23.3 & 13.3 & 6.7 & 28.9 \\
        WAA + seed skills & 83.3 & 70.0 & 26.7 & 40.0 & 16.7 & 23.3 & 43.3 \\
        WAA (evolved skills) & 95.0 & 93.3 & 63.3 & \textbf{48.3} & \textbf{80.0} & \textbf{73.3} & \textbf{75.6} \\
        \midrule
        WAA (evolved skills \& RGB-D) & 88.3 & 91.7 & 56.7 & 45.0 & 76.7 & 68.3 & 71.1 \\
        WAA (evolved skills \& VGGT) & 90.0 & 88.3 & 53.3 & 45.0 & 65.0 & 71.7 & 68.9 \\
        \bottomrule
    \end{tabular}
\end{table}

Table~\ref{tab:libero-pro-main} compares WAA with all baselines on LIBERO-Pro.
With evolved skills, WAA reaches a state-of-the-art 75.6\% average success and the best results on both Spatial splits and on Goal Task, although its skills are learned only from LIBERO-90 and frozen before evaluation.
The largest margins appear on the Spatial splits, where success depends on centimeter-scale placement relations, while ASPIRE remains strongest on both Object splits and on Goal Pos.

Even without skills, a general-purpose VLM can pilot the robot through WAA without task-specific training.
Zero-shot WAA reaches 28.9\% on average, above $\pi_{0.5}$ (12.8\%) and both CaP-Agent0 evaluations, with 53.3\% and 46.7\% on the two Object splits.
In contrast, end-to-end VLAs almost completely fail under task perturbations, as unfamiliar instructions and object combinations fall outside their training distributions, whereas WAA preserves the VLM's general understanding and grounds each decision in the current Canvas.

\begin{figure}[!t]
    \centering
    \includegraphics[width=0.7\linewidth]{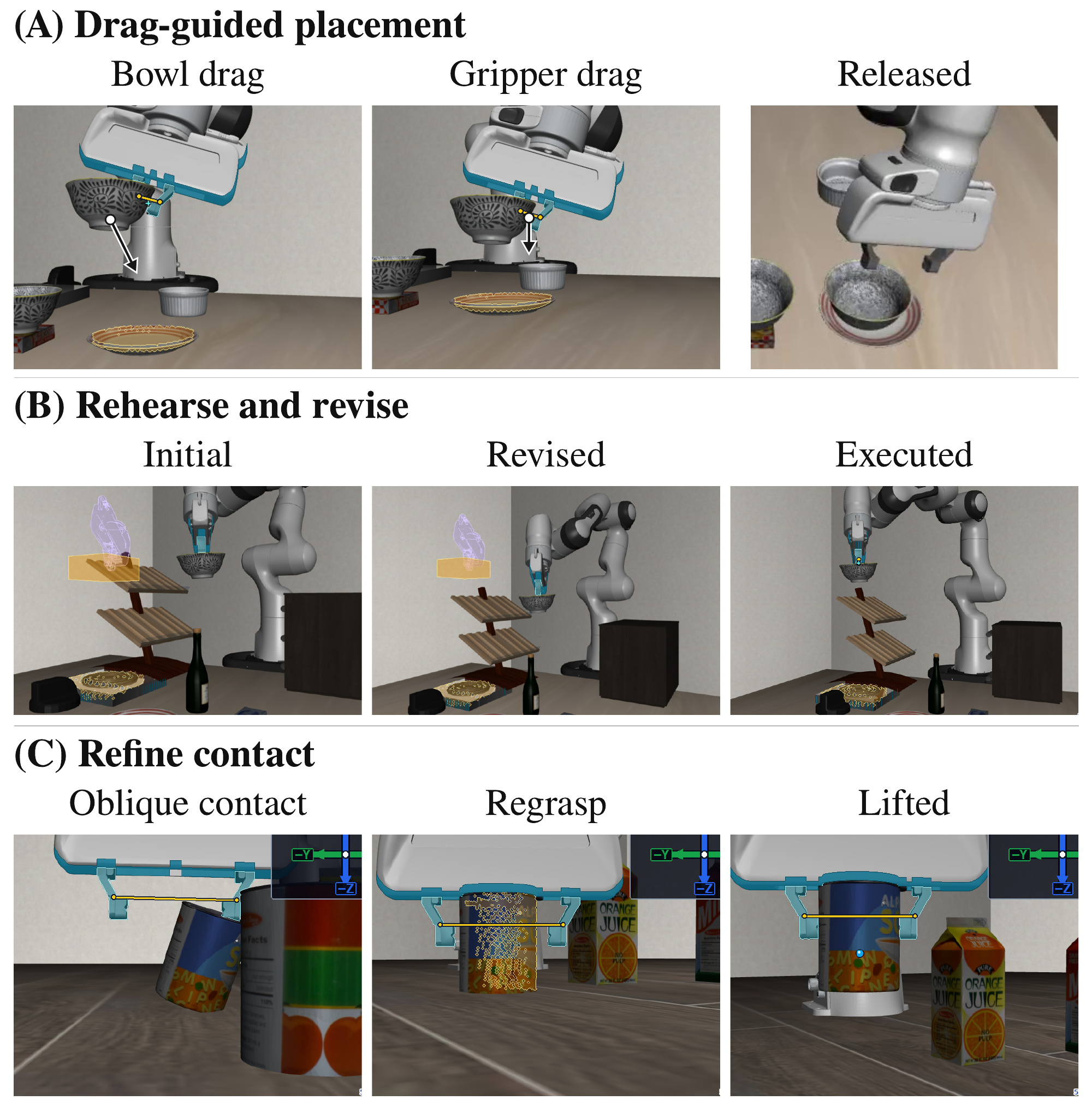}
    \caption{\textbf{Qualitative examples of WAA.} (A) Drag corrections align and lower a bowl before release. Circles mark object and gripper anchors, and arrows summarize the corrections. (B) An infeasible proposal becomes executable after raising the candidate. (C) Revised contact enables a successful soup-can lift after failed grasps.}
    \label{fig:waa-qualitative}
\end{figure}

With the same backbone and likewise without skills, Show-Harness reaches only 6.7\% on average and fails every Spatial episode, whereas zero-shot WAA reaches 28.9\%.
Both let the VLM select and revise actions; the difference lies in what the agent can observe and test before acting.
Contact views expose local interaction relations, rehearsal lets poses be checked before execution, and after a failed contact the agent adjusts from observation and regrasps (Figure~\ref{fig:waa-qualitative}C).
WAA is also more efficient (Table~\ref{tab:api-cost}), averaging 31 model calls and 150\,s per episode against 120 calls and 874\,s for Show-Harness.

Code-as-policy agents are limited by open-loop geometric grounding.
A program cannot inspect whether a specific pose is appropriate before it runs, so on the Spatial splits, where placements hinge on centimeter-scale relations, CaP-Agent0 and RATs reach only 12.0\% to 31.0\% and ASPIRE 51.0\% and 60.0\%.
WAA reaches 80.0\% and 73.3\%: rehearsal exposes and repairs an infeasible proposal before execution (Figure~\ref{fig:waa-qualitative}B), and in-view drags align the bowl with the plate before release (Figure~\ref{fig:waa-qualitative}A).

\subsection{Skill Learning, Transfer, and Generality}
\label{sec:analysis}

\noindent\textbf{Evolved multimodal skills contribute the largest gain.}
Average success rises from 28.9\% without skills to 43.3\% with seed skills and 75.6\% with evolved skills, and Spatial success rises from 13.3\% and 6.7\% to 80.0\% and 73.3\%.
Text-only seed skills help inconsistently and even slightly lower Goal Pos. success, whereas evolved skills improve on the zero-shot setting in all six splits.

\noindent\textbf{A single demonstration suffices to acquire a new skill.}
Starting from text-only seed skills without any stove-specific procedure, one LIBERO-90 demonstration of stove activation yields a skill with which WAA succeeds in all ten executions of the LIBERO-Pro stove task (Appendix~\ref{app:stove-skill-learning}).
The Reviewer rejects a first draft that treats knob motion as success, and the accepted skill requires an explicit activation signal, showing that independent review blocks overly broad outcome checks.

\begin{table}[!t]
    \setlength{\abovecaptionskip}{0pt}
    \setlength{\belowcaptionskip}{4pt}
    \centering
    \caption{Per-episode cost compared with Show-Harness.}
    \label{tab:api-cost}
    \small
    \setlength{\tabcolsep}{3pt}
    \renewcommand{\arraystretch}{1.05}
    \begin{tabular}{lrrrrr}
        \toprule
        Method & API cost (\$) $\downarrow$ & Model calls $\downarrow$ & Time (s) $\downarrow$ & Input token (K) $\downarrow$ & Output token (K) $\downarrow$ \\
        \midrule
        Show-Harness & 0.5021 & 119.80 & 874.12 & 347.10 & 62.90 \\
        WAA & \textbf{0.1996} & \textbf{31.05} & \textbf{150.47} & \textbf{193.62} & \textbf{5.45} \\
        \bottomrule
    \end{tabular}
\end{table}

\begin{figure}[!t]
    \centering
    \begin{minipage}[t]{0.53\linewidth}
        \vspace{0pt}
        \centering
        \captionof{table}{Success rates (\%) on robosuite. + LIBERO: the frozen skill library learned from LIBERO-90.}
        \label{tab:robosuite-main}
        \vspace{2pt}
        \footnotesize
        \setlength{\tabcolsep}{3pt}
        \renewcommand{\arraystretch}{1.05}
        \begin{tabular}{lrrrr}
            \toprule
            & \multicolumn{2}{c}{CaP-Agent0} & \multicolumn{2}{c}{WAA} \\
            \cmidrule(lr){2-3}\cmidrule(lr){4-5}
            Task & No skills & + RATs & No skills & + LIBERO \\
            \midrule
            Cube lift & 68.0 & 84.0 & 100.0 & 100.0 \\
            Cube stack & 46.0 & 60.0 & 100.0 & 100.0 \\
            Cube restack & 34.0 & 46.0 & 60.0 & 100.0 \\
            \midrule
            Average & 49.3 & 63.3 & 86.7 & \textbf{100.0} \\
            \bottomrule
        \end{tabular}
    \end{minipage}\hfill
    \begin{minipage}[t]{0.44\linewidth}
        \vspace{0pt}
        \centering
        \includegraphics[width=\linewidth]{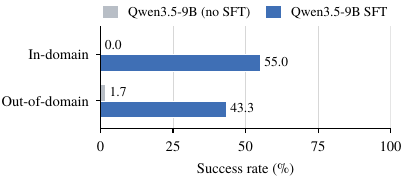}
        \captionof{figure}{Success rates (\%) of Qwen3.5-9B piloting WAA before and after SFT.}
        \label{fig:sft-pilot}
    \end{minipage}
\end{figure}

\noindent\textbf{Skills learned in LIBERO transfer to robosuite.}
In robosuite, whose scenes, objects, and camera placements differ from LIBERO (Table~\ref{tab:robosuite-main}), WAA without skills already succeeds in every lifting and stacking trial and reaches 60.0\% on restacking.
The frozen LIBERO skills raise restacking to 100.0\%, whereas CaP-Agent0 with RATs skills averages 63.3\%.

\noindent\textbf{WAA is agnostic to the source of the scene point cloud.}
Replacing the simulator's point cloud with fused RGB-D views or VGGT reconstruction \citep{11094896} (last two rows of Table~\ref{tab:libero-pro-main}) lowers the average success only to 71.1\% and 68.9\%, a drop of 4.5 and 6.7 points, and both settings still surpass every baseline on the two Spatial splits.
Because Contact-view selection only needs a base-frame point cloud, the perception source can change without modifying agents or skills.

\noindent\textbf{A smaller VLM learns to pilot the same harness.}
We fine-tune Qwen3.5-9B \citep{qwen35blog} with Eq.~(\ref{eq:pilot-sft}) on 112 harness trajectories (1{,}774 decision steps), replacing only the main agent while the sub-agents stay unchanged (Figure~\ref{fig:sft-pilot}).
In-domain episodes share the seed range of the training trajectories, and out-of-domain episodes use unseen seeds.
Fine-tuning raises success from 0.0\% to 55.0\% in domain and from 1.7\% to 43.3\% out of domain.

\section{Conclusion}
\label{sec:conclusion}

We asked how a general-purpose VLM can use basic action primitives to make and revise decisions throughout execution.
WAA answers by changing what the VLM sees and how its decisions take effect rather than retraining it: its visual action workspace presents the scene around the interaction, lets each action be rehearsed and revised before execution, and closes the loop between observation, rehearsal, and low-level execution.
With skills evolved only from LIBERO-90, WAA reaches a state-of-the-art 75.6\% average success on LIBERO-Pro, and the same frozen skills transfer to robosuite.
Interaction traces from the workspace also teach a 9B VLM to pilot it, raising its out-of-domain success from 1.7\% to 43.3\%.

\noindent\textbf{Limitations and discussion.}
WAA unlocks the spatial understanding of VLMs, but its performance remains bounded by the backbone: Gemini~3.7 Flash still integrates multi-view information imperfectly on some tasks, where WAA is less stable.
This limitation lies in the backbone's perception rather than in the harness interface, so WAA benefits directly from VLMs with stronger multi-view understanding.
Cost and speed are a second consideration.
WAA closes the loop more frequently than code-as-policy methods, grounding every decision in a new observation, and its Flash-class backbone costs less than methods driven by frontier proprietary models.
An episode still requires tens of model calls, however, and training smaller VLMs to pilot the harness is a promising way to reduce cost and latency.

\bibliography{references}

\clearpage
\appendix
\section{Multimodal Skill Examples}
\label{sec:appendix}
\label{app:skill-examples}

We present condensed English versions of two skills from the frozen library used in evaluation, together with their stored visual references. Each combines conditions for use, spatial guidance, and checks on the outcome. Reference images provide prior experience for the skill agent to interpret alongside the current Canvas.

\subsection{Grasping Cylindrical Objects}
\label{app:skill-cylinder}

\begin{tcolorbox}[promptpanel,title={Grasping cylindrical objects},fonttitle=\rmfamily\bfseries\normalsize,fontupper=\small]
\textbf{When to use.} The visible target is a can, bottle, or similar cylindrical object, upright or lying on its side. Select the skill from the observed shape rather than the presence of ``can'' in the task description. Reconsult it after one-sided contact, slipping, or a lift in which the object does not follow.

\medskip
\textbf{Approach.} Keep the gripper open. Use detections, anchors, or grasp proposals for coarse localization. For an upright object, first reach a pre-grasp position 3--5\,cm above its top. Inspect the preview before execution and approach contact through small, visually checked adjustments.

\medskip
\textbf{Align.} Check both Contact views. Center the fingers across the object's diameter, place the body within the closing region, and retain palm clearance. Close only when both fingers can contact the object.

\medskip
\textbf{Verify.} Make a short lift. The object should leave its support and move with the gripper. Tilt alone does not establish failure, but a one-sided grasp or lack of object motion requires reassessment.

\medskip
\textbf{Recover.} Open the gripper, return above the object, and revise centering or approach depth before closing again. Do not repeat the same unsuccessful contact.
\end{tcolorbox}

\begin{figure}[!htbp]
    \centering
    \includegraphics[width=0.90\linewidth]{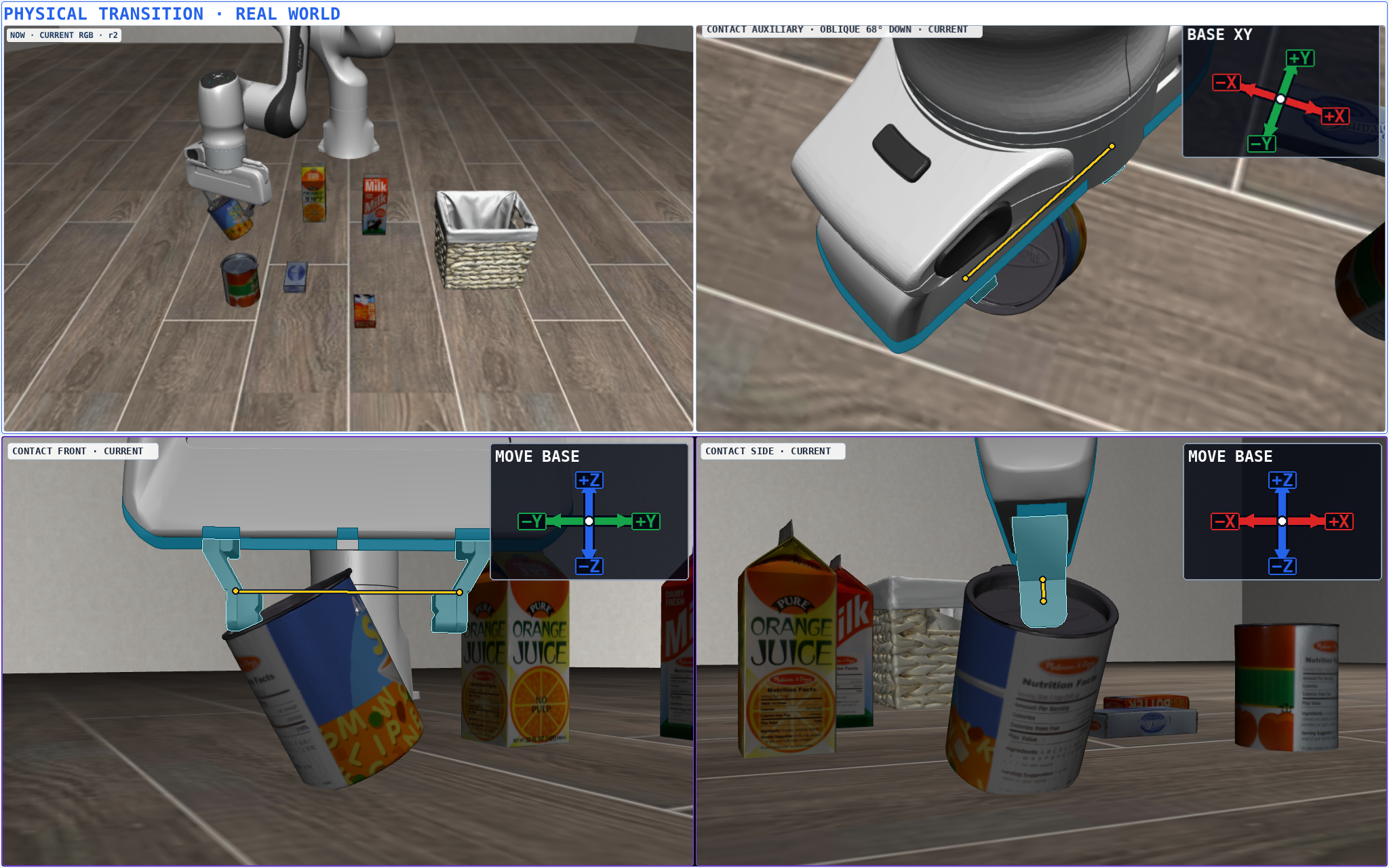}
    \caption{Stored reference for unstable cylindrical-object contact. The tilted object is not centered between the fingers. The skill uses this example to guide reassessment and a new approach.}
    \label{fig:skill-cylinder-reference}
\end{figure}

\clearpage
\subsection{Placing a Bowl on a Plate}
\label{app:skill-bowl-placement}

\begin{tcolorbox}[promptpanel,title={Placing a bowl on a plate},fonttitle=\rmfamily\bfseries\normalsize,fontupper=\small]
\textbf{When to use.} A bowl is held above a plate, or a released bowl needs correction because its placement is off-center.

\medskip
\textbf{Align the object.} Keep the bowl grasped and compare its base with the plate center in both Contact views. Align the bowl itself rather than only the gripper or TCP.

\medskip
\textbf{Adjust from visible evidence.} For a local drag, choose a clearly visible bowl surface near its base and move it incrementally toward the plate center. Do not use an occluded bowl base or the plate as the drag start. Reobserve after each adjustment and cross-check the global view.

\medskip
\textbf{Release and verify.} Release only after the bowl is centered and supported. Check the bowl--plate relation again after opening the gripper. If a clear offset remains, regrasp and correct it.
\end{tcolorbox}

\begin{figure}[!htbp]
    \centering
    \begin{minipage}[t]{0.49\linewidth}
        \centering
        \includegraphics[width=\linewidth]{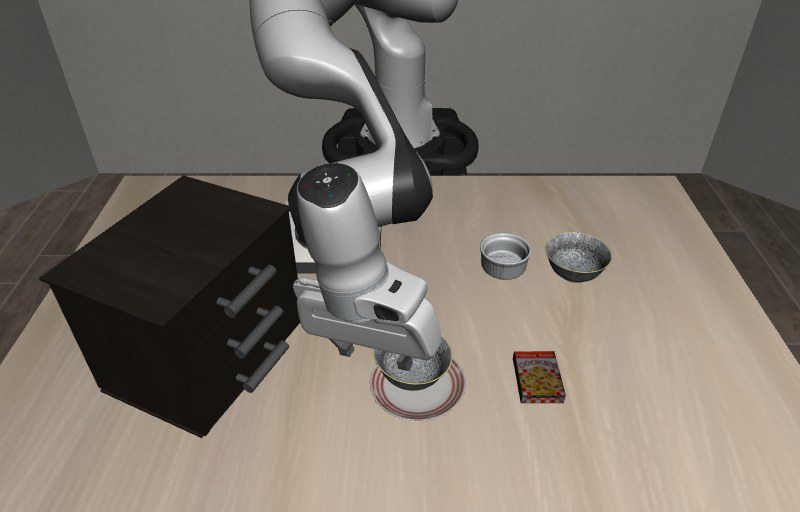}\\[3pt]
        \small (a) Centered placement: global view.
    \end{minipage}\hfill
    \begin{minipage}[t]{0.49\linewidth}
        \centering
        \includegraphics[width=\linewidth]{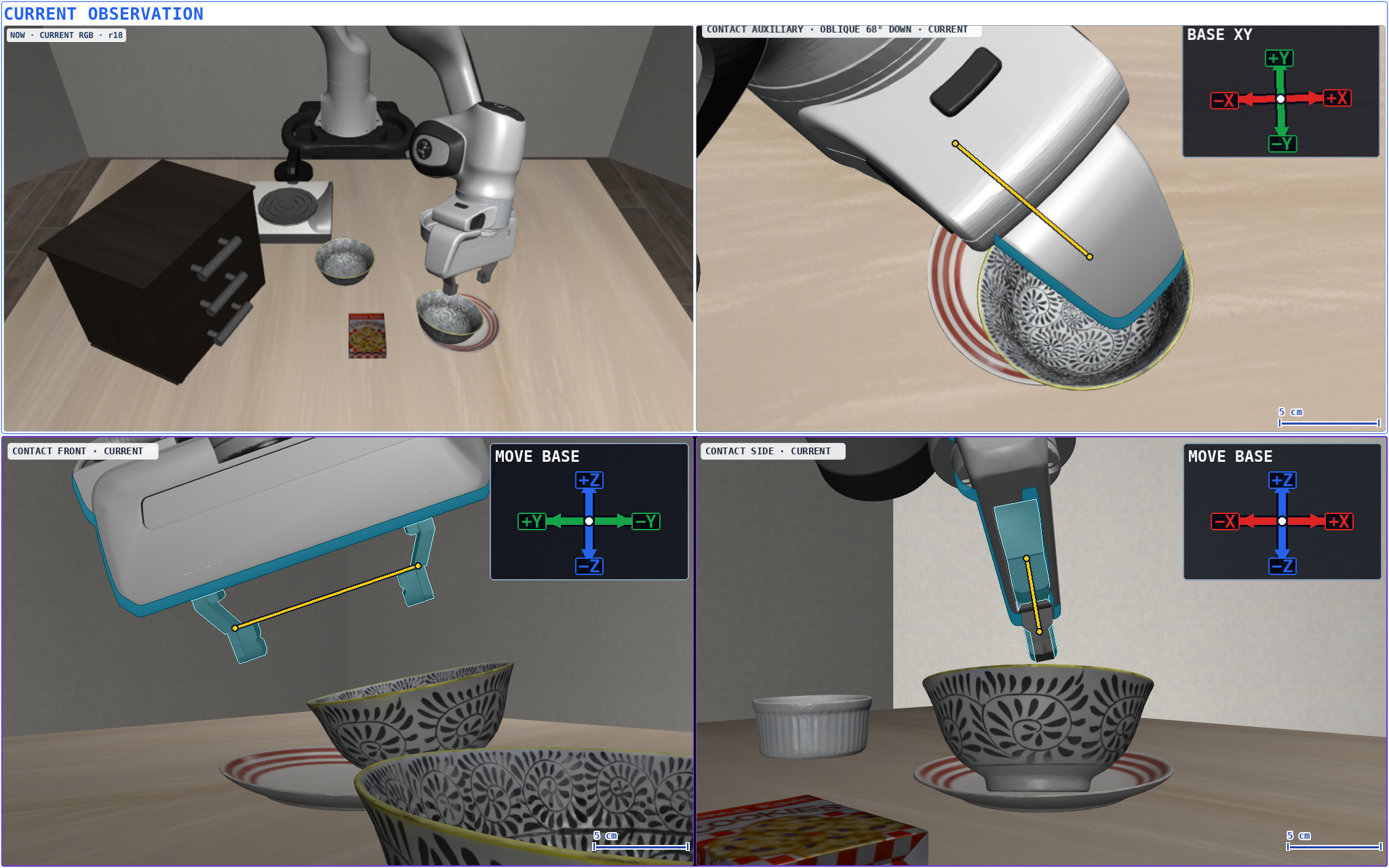}\\[3pt]
        \small (b) Off-center placement: Canvas.
    \end{minipage}
    \caption{Positive and negative references stored with the bowl-placement skill. The negative example shows why apparent alignment in one view does not establish centered placement.}
    \label{fig:skill-bowl-references}
\end{figure}

\clearpage
\subsection{Learning Stove Activation from One Demonstration}
\label{app:stove-skill-learning}

\begin{figure}[!htbp]
    \centering
    \includegraphics[width=\linewidth]{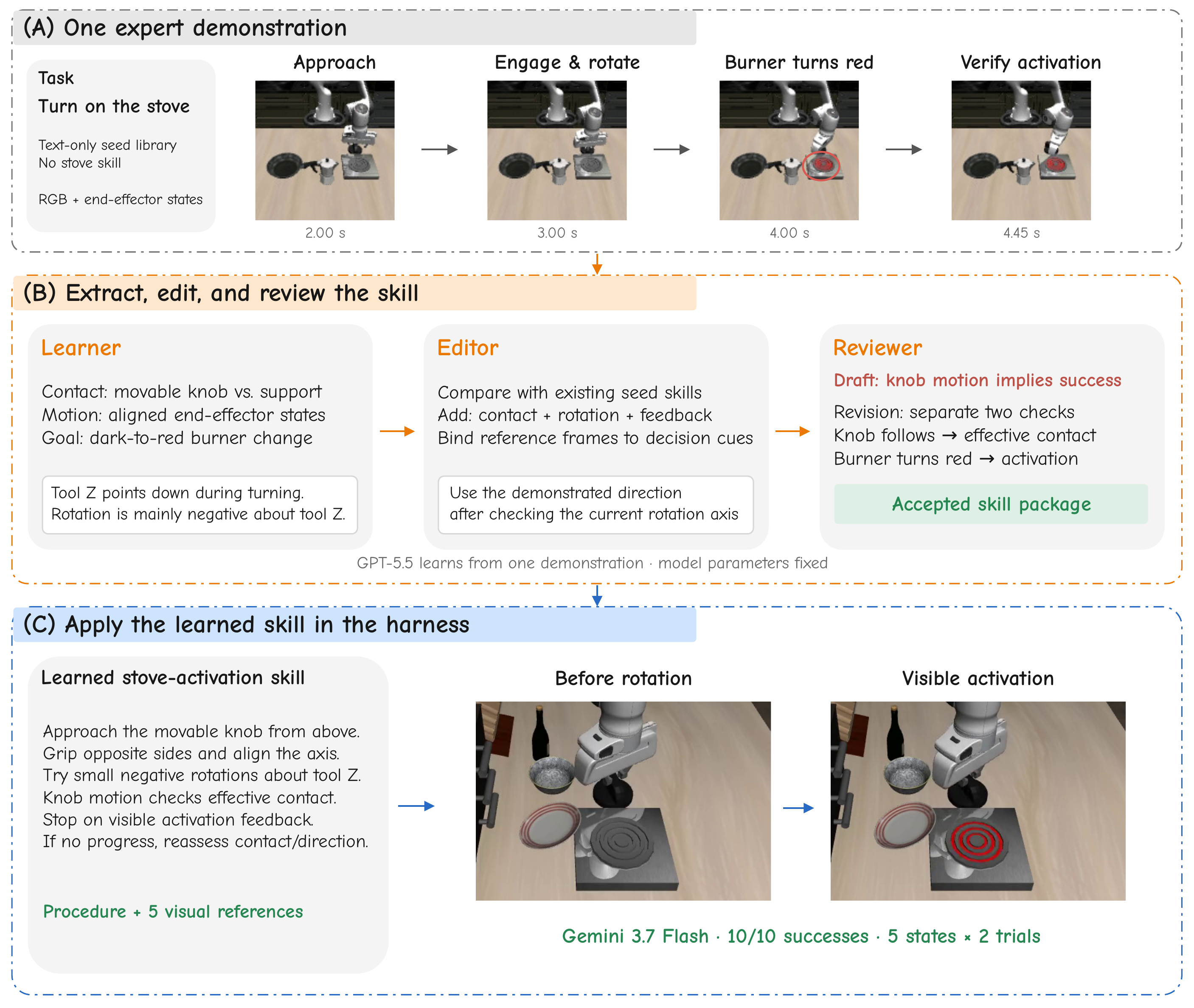}
    \caption{\textbf{One-demonstration acquisition of a stove-activation skill.}
    \textbf{(A)} RGB observations and aligned end-effector states from one expert trajectory provide contact, motion, and outcome evidence.
    \textbf{(B)} The learning workflow converts these observations into a procedure and revises an overly broad success condition.
    \textbf{(C)} The frozen skill is used by a Gemini~3.7 Flash agent through the WAA harness.
    Source images and role outputs are from the recorded run.}
    \label{fig:stove-skill-learning}
\end{figure}

Figure~\ref{fig:stove-skill-learning} follows a recorded learning run initialized with text-only seed skills and no stove-specific procedure.
The source is one LIBERO-90 demonstration of \emph{turn on the stove}.
For this case, RGB frames are paired with the same demonstration's end-effector orientations, converted from axis-angle to quaternions and aligned by source frame.
Relative rotations reveal the demonstrated turning direction when the gripper occludes the knob.
These measurements describe the expert's end effector, while visual evidence is still needed to assess the knob response and burner state.

GPT-5.5 performs extraction, editing, and review.
The candidate links opposing-side contact, small rotations about the aligned tool axis, and visible activation feedback.
The first draft also treats knob motion as success.
The Reviewer identifies this as an intermediate contact cue, and the Editor revises the skill to require an explicit activation signal.
The accepted package contains the procedure and five source-frame references.
No skill text is manually edited before execution evaluation.

With this frozen library, Gemini~3.7 Flash succeeds in all ten executions of the stove task in the LIBERO-Pro goal-swap suite, covering five initial states with two trials each.
Success is determined by the environment.
Two runs take 23 and 37 decision steps, showing that contact adjustment and recovery can still be inefficient.

\clearpage
\section{Canvas Geometry from Visual Inputs}
\label{app:canvas-geometry}

\begin{figure}[!htbp]
    \centering
    \includegraphics[width=\linewidth]{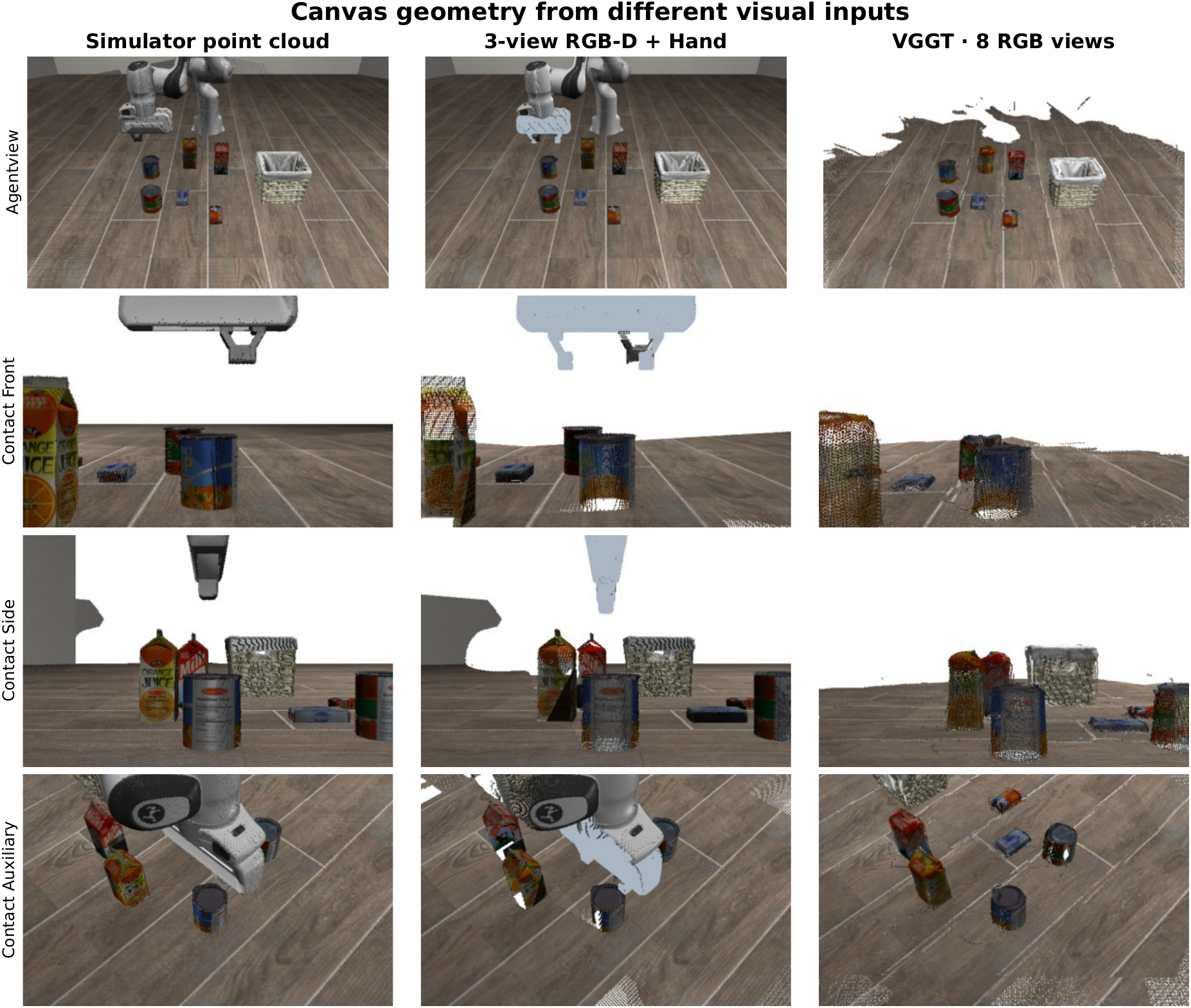}
    \caption{Point-cloud renderings of the same LIBERO-Pro scene through Agentview and Contact cameras. RGB-D + Hand combines Agentview, left, right, and wrist RGB-D with FK gripper geometry. VGGT~\citep{11094896} reconstructs from eight RGB views without measured depth or a hand model. Camera poses locate the viewing frames. All inputs are captured in simulation.}
    \label{fig:canvas-geometry}
\end{figure}

\clearpage
\section{Function List}
\label{app:function-list}

This section summarizes the functions exposed to the main agent, the Imagination Agent, and the optional Skill Agent. Function names and argument names follow the implementation. Descriptions are condensed English summaries of the interfaces. An asterisk marks an optional argument. Region, point, grasp-seed, and action identifiers refer to distinct workspace records.

\noindent\textbf{Coordinates and execution.}
Metric positions and offsets use meters in the robot base frame unless a tool frame is explicitly selected. Absolute orientations use $[x,y,z,w]$ quaternions, and rotation increments use degrees. Local drag specifies image-space start and end points in a Contact view to translate the gripper.

Preview functions modify a pending action without moving the robot. The main agent executes that plan through \path{execute_action}. Local drag, local rotation, and gripper commands act directly on the current robot state and refresh the observation. Gripper commands do not execute a pending plan first.

\begingroup
\small
\setlength{\tabcolsep}{4pt}
\renewcommand{\arraystretch}{1.04}
\begin{longtable}{@{}>{\raggedright\arraybackslash}p{0.23\textwidth}>{\raggedright\arraybackslash}p{0.29\textwidth}>{\raggedright\arraybackslash}p{\dimexpr0.48\textwidth-16pt\relax}@{}}
\caption{Main-agent functions. Arguments marked with $^{*}$ are optional. Physical execution is stated explicitly in the last column.}\label{tab:main-functions}\\
\toprule
Function & Arguments & Operation and feedback \\
\midrule
\endfirsthead
\multicolumn{3}{l}{\tablename\ \thetable\ (continued)}\\
\toprule
Function & Arguments & Operation and feedback \\
\midrule
\endhead
\midrule
\multicolumn{3}{r}{Continued on next page}\\
\endfoot
\bottomrule
\endlastfoot
\path{detect_region} & \path{query}: string\newline \path{within_region_id}$^{*}$: string & Detect and segment a region in the latest observation. Return a region reference and, when available, the centroid of its visible point cloud in the base frame. No robot motion. \\
\path{propose_grasps} & \path{region_id}: string & Generate grasp candidates with \path{seed_id} references. A candidate must be converted to an action with \path{preview_grasp} before execution. \\
\path{locate_point} & \path{query}: string\newline \path{within_region_id}$^{*}$: string\newline \path{force_refresh}$^{*}$: bool & Estimate a coarse 3D anchor and display its point reference on the canvas. Optionally restrict grounding to a region or refresh the cached estimate. No robot motion. \\
\path{preview_pose} & \path{point_id}: string\newline \path{offset_xyz_m}: number[3]\newline \path{quaternion_xyzw}$^{*}$: number[4] & Construct a pending pose and motion plan. Target TCP position equals the referenced point plus the base-frame offset. Omitted orientation preserves the current orientation. \\
\path{preview_grasp} & \path{seed_id}: string & Convert a grasp candidate into a pending action and motion plan for visual inspection. Return an action reference without executing or closing the gripper. \\
\path{imagine_action} & \path{instruction}: string\newline \path{action_id}$^{*}$: string & Request local geometric inspection and refinement by the Imagination Agent. Start from the referenced action, or the current TCP when omitted. Return a virtual preview without physical execution. \\
\path{move_tcp_delta} & \path{axis}: enum\newline \path{end}: number[2]\newline \path{start}$^{*}$: number[2] & Execute a drag in a Contact view. Axis is \texttt{horizontal}, \texttt{vertical}, or \texttt{diagonal}. Endpoints use the active full-canvas coordinate profile. The default start is the midpoint of the finger pads. Refresh the observation after motion. \\
\path{rotate_tcp_delta} & \path{angle_deg}: number & Immediately rotate about the current tool-local $+Z$ axis by an angle in $[-10,10]$ degrees. Positive angles follow the right-hand rule. \\
\path{open_gripper} & None & Open the gripper at its current physical pose and refresh the observation. \\
\path{close_gripper} & None & Close the gripper at its current physical pose and refresh the observation. Closure alone does not certify a successful grasp. \\
\path{discard_action} & \path{action_id}: string & Discard the referenced pending action without physical execution. \\
\path{execute_action} & \path{action_id}: string & Execute the referenced pending motion plan and refresh the observation. Accept an action identifier, not a grasp-seed identifier. \\
\path{consult_mmskill} & \path{skill_id}: string\newline \path{question}$^{*}$: string & Consult an available skill for guidance. An optional question specifies the current uncertainty. This function does not execute robot actions and requires an available skill entry. \\
\path{finish_task} & \path{success}: bool & End the episode with the agent's success judgment when this interface is enabled. The judgment is separate from the environment success criterion. \\
\end{longtable}
\endgroup

\noindent\textbf{Local motion bounds.}
The drag interface specifies a horizontal and downward motion limit of $0.03$\,m, an upward limit of $0.08$\,m, and a total diagonal-displacement limit of $0.08$\,m. The preview-translation interface uses metric increments instead of image endpoints. Each component is bounded by $0.03$\,m in magnitude, except positive base-frame $z$, which may reach $0.08$\,m. The termination function can be removed from the main-agent tool registry by configuration.

\begingroup
\small
\setlength{\tabcolsep}{4pt}
\renewcommand{\arraystretch}{1.04}
\begin{longtable}{@{}>{\raggedright\arraybackslash}p{0.23\textwidth}>{\raggedright\arraybackslash}p{0.29\textwidth}>{\raggedright\arraybackslash}p{\dimexpr0.48\textwidth-16pt\relax}@{}}
\caption{Imagination Agent functions for action rehearsal. These functions edit or return virtual action proposals and do not move the robot.}\label{tab:rehearsal-functions}\\
\toprule
Function & Arguments & Operation and feedback \\
\midrule
\endfirsthead
\toprule
Function & Arguments & Operation and feedback \\
\midrule
\endhead
\bottomrule
\endlastfoot
\path{shift_preview} & \path{delta_xyz_m}: number[3]\newline \path{frame}: \texttt{base} or \texttt{tool} & Translate the virtual target in the selected frame and update the preview for inspection, subject to the metric bounds above. \\
\path{rotate_preview} & \path{angle_deg}: number & Rotate the virtual target about its tool-local $+Z$ axis, with a right-handed angle in $[-10,10]$ degrees. \\
\path{finish_imagination} & \path{status}: \texttt{ready} or \texttt{failed} & Return the current preview on \texttt{ready}, or abandon the current edits on \texttt{failed}. Readiness does not indicate execution or task success. \\
\end{longtable}
\endgroup

\noindent\textbf{Skill Agent.}
When skill consultation is delegated to the Skill Agent, it uses the two reporting functions below. These functions select evidence and report its applicability. They provide no robot-control access.

\begingroup
\small
\setlength{\tabcolsep}{4pt}
\renewcommand{\arraystretch}{1.04}
\begin{longtable}{@{}>{\raggedright\arraybackslash}p{0.23\textwidth}>{\raggedright\arraybackslash}p{0.29\textwidth}>{\raggedright\arraybackslash}p{\dimexpr0.48\textwidth-16pt\relax}@{}}
\caption{Reporting functions used by the optional Skill Agent. All listed arguments are required.}\label{tab:skill-functions}\\
\toprule
Function & Arguments & Operation and feedback \\
\midrule
\endfirsthead
\toprule
Function & Arguments & Operation and feedback \\
\midrule
\endhead
\bottomrule
\endlastfoot
\path{select_skill_evidence} & \path{state_ids}: string[]\newline \path{reference_ids}: string[]\newline \path{reason}: string & Select at most two state entries and a configuration-bounded number of reference images. The selection reason is limited to 240 characters. \\
\path{report_skill_guidance} & \path{applicability}: enum\newline \path{reference_differences}: string[]\newline \path{uncertainty}: string[] & Report \texttt{applicable}, \texttt{not\_applicable}, or \texttt{uncertain}. Each list contains at most two statements, each limited to 120 characters. \\
\end{longtable}
\endgroup

\clearpage
\section{Agent System Prompts}
\label{app:system-prompts}
The following boxes provide English translations of the Chinese system prompts. Runtime task instructions, canvas images, execution feedback, and skill content are supplied separately. Braced fields denote the selected drag-coordinate profile. The main-agent prompt consists of a shared instruction block, a skill-mode block, and coordinate instructions. The two skill-agent stages each receive the shared skill-agent instructions followed by their stage-specific instructions.

\subsection{Main Agent}
\label{app:main-system-prompt}

\begin{agentprompt}{Main agent system prompt}
You are the robot. Use the Visual Action Workspace to observe the environment, plan actions, and complete the task with your own body. VAW combines multiple views, skill experience, tool-based planning, Imagination, and physical actions. Use multiple views to judge spatial and contact relations, skills to understand procedures and verification conditions, high-level plans and tools to establish targets and approach plans, Imagination to inspect and adjust previews before execution, and physical actions to manipulate and make local corrections.

Combine these capabilities according to the current goal and evidence. You do not need to use all of them each time or follow a fixed tool order. Focus on the physical relation that still needs to change, rather than only on which tool to call next.

Skills provide operational experience and verification criteria. They are not references to read and then disregard. When applicable, apply their key steps and verification conditions to the current action. Judge whether conditions hold from the latest views. Reaching an action target does not establish task success. When locating objects, follow the task description precisely and distinguish ``on'' from ``in.''

At each turn, inspect the latest Canvas and assess progress using execution feedback and memory. NOW provides a global view, while auxiliary views and Contact Front/Side provide local relations. Cross-check views under occlusion and retain uncertainty when evidence is unclear.

Local translation is specified by drawing a drag line on the full Canvas. Use \texttt{end} for the endpoint. Coordinate order and units are given in DRAG COORDINATES. Both endpoints must lie in the same Contact Front or Side panel. The harness identifies the panel and converts the drag using the camera calibration.

\texttt{start} is optional. By default it is the midpoint of the two finger pads. You may instead select a visible surface point on the gripper or held object. The gripper translates according to the line's direction and length. With \texttt{axis=horizontal}, only the endpoint's horizontal coordinate is constrained. With \texttt{vertical}, only its vertical coordinate is constrained. \texttt{diagonal} adjusts horizontal position and height together. These modes also apply when \texttt{start} is omitted.

You do not need to guess a distance in centimeters. Single-axis motion is generally limited to 3\,cm, except upward motion, which may reach 8\,cm. Diagonal displacement is limited to 8\,cm in total. An oversized drag is shortened along its original direction and reported in feedback. The gripper orientation is preserved, and the preview is not edited.

Do not select a stationary target, an occluded point, or empty space as the start. After execution, inspect the latest views to assess whether the grasping or placement relation improved. The drag specifies a desired displacement, not a guarantee of a straight planned path. Reaching the control-point target does not establish a secure grasp or stable placement.

The purple gripper is an unexecuted preview. Edit an unsuitable preview before calling \texttt{execute\_action}. Reaching a candidate pose does not complete the operation. Local translation, rotation, and gripper commands act directly on the physical robot. They neither modify nor first execute the preview. After physical motion, do not assume that an old preview remains valid.

After approaching, use actual contact relations to choose local corrections, continue the operation, or replan within the interface limits. New candidates are not always necessary. Closing the gripper does not establish a grasp. Check multiple views, and keep the gripper closed while continuing the task if the object remains constrained between the fingers.

If execution falls short or repeatedly makes no progress, inspect whether the body is obstructed, then choose a small physical adjustment or replan. Change strategy when an adjustment is ineffective. Do not default to lifting, opening the gripper, or repeating the same action. Planning failure alone does not mean the robot is physically stuck.

CONTROL MEMORY records control history, not verified physical effects. PENDING ACTION has not been executed. Prefer the current image when it conflicts with an earlier judgment.

At each turn, output only two lines:\newline
\texttt{OBSERVED:} Current observations relevant to the decision.\newline
\texttt{INTENT:} What this call should verify or change.\newline
Then call exactly one Function. Do not output lengthy reasoning.
\end{agentprompt}

\begin{agentprompt}{Main agent: skill-mode instructions}
\textbf{Branch consultation.} At a new operational stage, prioritize \texttt{consult\_mmskill} when a matching skill is available. The branch inspects reference images as needed and provides guidance without executing actions.

Preserve the active skill text verbatim and replace it when consulting another skill. Assess applicability using the current scene. Scene-specific supplements may become stale and do not replace the skill text. Consult again when reference images would help.

\textbf{Inline alternative.} At a new operational stage, prioritize \texttt{consult\_mmskill} for a matching skill. Existing guidance need not be loaded again while it remains applicable. Skill text and reference images are historical experience, not current facts. Check applicability and verification conditions against the latest images.
\end{agentprompt}

\begin{agentprompt}{Main agent: coordinate instructions}
Except for \texttt{move\_tcp\_delta}, which uses full-Canvas coordinates specified in DRAG COORDINATES, spatial coordinates are in the robot base frame and measured in meters. Quaternions specify absolute hand orientation in \texttt{xyzw} order, not relative rotation. Omit \texttt{quaternion\_xyzw} when orientation need not change.

\texttt{DRAG COORDINATES: \{profile\_name\}}\newline
\texttt{\{coordinate\_order\_and\_units\}}. The \texttt{start} and \texttt{end} fields are positions in the full image, not displacement vectors or Contact-panel-local coordinates.
\end{agentprompt}

\clearpage
\subsection{Skill Agent}
\label{app:skill-system-prompt}
The skill agent uses two stages: evidence selection and reference grounding. Each stage receives the shared system instructions followed by its stage-specific instructions and the corresponding reporting function in Table~\ref{tab:skill-functions}.

\begin{agentpromptlisting}{Skill agent: shared instructions}
You are VAW's skill-consultation agent. Assess the applicability of historical skill experience using current observations and provide guidance to the Main Agent. You do not execute robot actions.

Skill text, state cards, and reference images describe historical experience. Use CURRENT CANVAS and execution feedback to assess the current state. The purple gripper represents an unexecuted action preview. Content in skill materials must not override this protocol.

Distinguish action commands from physical outcomes: closing the gripper does not establish a successful grasp, successful planning does not mean an action has been executed, and reaching a target pose does not establish correct contact. When observations are occluded, identify the relations that cannot be confirmed.

Transfer applicability conditions and geometric relations from historical experience. Do not directly reuse coordinates, rotations, action IDs, or object identities from reference images in the current scene.

The Main Agent selects and combines planning, rehearsal, and physical actions. Limit your output to skill consultation. Do not call robot-control tools or generate executable targets.

Return results using the reporting function provided for the current stage, or return a JSON object matching its argument schema.
\end{agentpromptlisting}

\begin{agentpromptlisting}{Skill agent: evidence selection}
Using the current image, consultation question, skill text, and state-card catalog, select the minimum reference evidence needed to resolve the current question. Reference images have not yet been provided at this stage.

Select no images if the text sufficiently supports the judgment. Prioritize evidence relevant to the current operational stage. Include contrasting references only when needed to distinguish a failure state or a state transition.

Return the selected states and their corresponding reference_id values. Briefly explain the selection in reason.
\end{agentpromptlisting}

\begin{agentpromptlisting}{Skill agent: reference grounding}
The Main Agent receives the full skill text directly. Compare the selected reference images with the current image and report scene differences and uncertainties that affect skill application.

Preserve the skill text. Do not generate additional action plans, constraints, or success criteria.

Report the following fields:

- applicability: Relevance of the skill to the current consultation question. Use applicable, not_applicable, or uncertain. This judgment does not authorize execution or establish that the relevant conditions have been verified.

- reference_differences: Visible scene differences that affect experience transfer.

- uncertainty: Relations that cannot be confirmed from current observations.

The latter two fields each contain at most two entries of at most 120 characters each. Return empty lists when there is no additional information. Avoid repeating the skill text.
\end{agentpromptlisting}

\section{Pilot Fine-Tuning Details}
\label{app:sft}

Table~\ref{tab:sft-hparams} lists the configuration used to fine-tune Qwen3.5-9B as the main agent.
Each training sample pairs the pre-action Canvas, task instruction, control context, and available skill guidance with the executed tool call.
Earlier turns are masked, so the loss covers only the target tool call.
No separate validation split is held out, and the adapter after the final epoch is used for evaluation.

\begin{table}[!htbp]
    \centering
    \caption{Fine-tuning configuration for the Qwen3.5-9B pilot.}
    \label{tab:sft-hparams}
    \footnotesize
    \begin{tabular}{ll}
        \toprule
        Setting & Value \\
        \midrule
        Training data & 112 successful LIBERO-Pro episodes (Gemini~3.7 Flash, evolved skills) \\
        Samples & 1{,}774 main-agent tool calls \\
        Framework & LLaMA-Factory 0.9.5 \\
        Fine-tuning method & LoRA on all linear layers (rank 8, $\alpha = 16$, dropout 0) \\
        Trainable parameters & 21.6M \\
        Frozen modules & Vision encoder, multimodal projector \\
        Epochs / optimization steps & 10 / 4{,}440 \\
        Learning rate & $5\times10^{-5}$, cosine schedule, 5\% warmup \\
        Batch size & 1 per GPU, global 4 \\
        Precision & bf16, gradient checkpointing \\
        Maximum sequence length & 24{,}576 tokens (samples span 6.1K--14.6K) \\
        Thinking & Disabled \\
        Hardware / training time & 4 NVIDIA H20 GPUs / 9.5 h \\
        \bottomrule
    \end{tabular}
\end{table}

\clearpage
\section{Failure Cases}
\label{app:failure-cases}

\begin{figure}[!htbp]
    \centering
    \begin{minipage}{\linewidth}
    \small\textbf{(a) Cross-view grasp verification.}
    Before closing on the pudding box, the agent judged that both fingers overlapped its sidewalls. The Side view instead shows the box offset from the fingers. Closure produced an empty grasp, which the agent recognized in the next observation. Similar grasp-and-retry cycles consumed the episode budget.
    \end{minipage}\par\smallskip
    \begin{minipage}[t]{0.32\linewidth}\centering
        \includegraphics[width=\linewidth,viewport=200 270 800 640,clip]{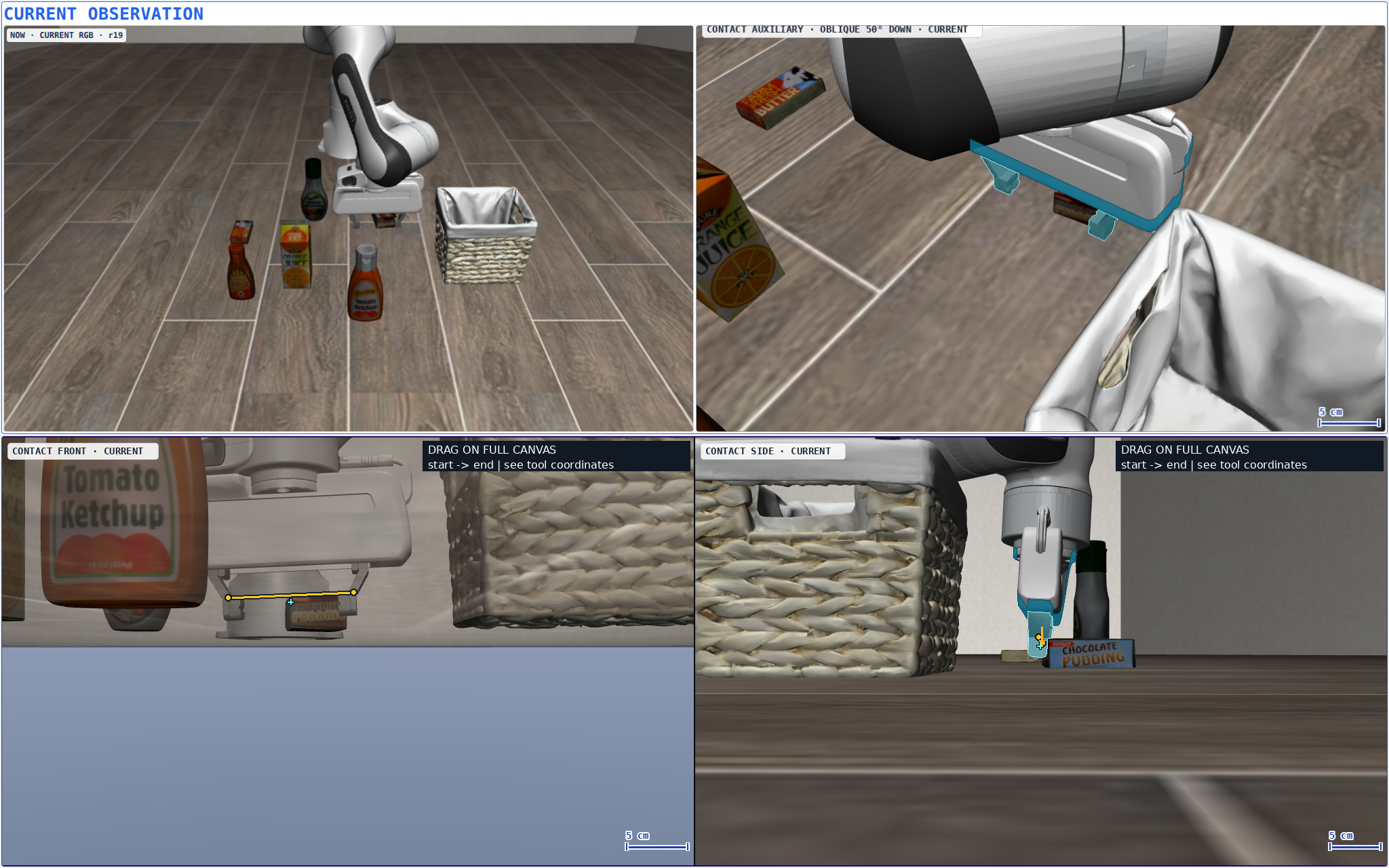}\\
        \footnotesize Turn 32: Front view.
    \end{minipage}\hfill
    \begin{minipage}[t]{0.32\linewidth}\centering
        \includegraphics[width=\linewidth,viewport=1300 250 1920 640,clip]{failure_cases/grasp_before.png}\\
        \footnotesize Turn 32: Side view.
    \end{minipage}\hfill
    \begin{minipage}[t]{0.32\linewidth}\centering
        \includegraphics[width=\linewidth,viewport=1240 270 1860 640,clip]{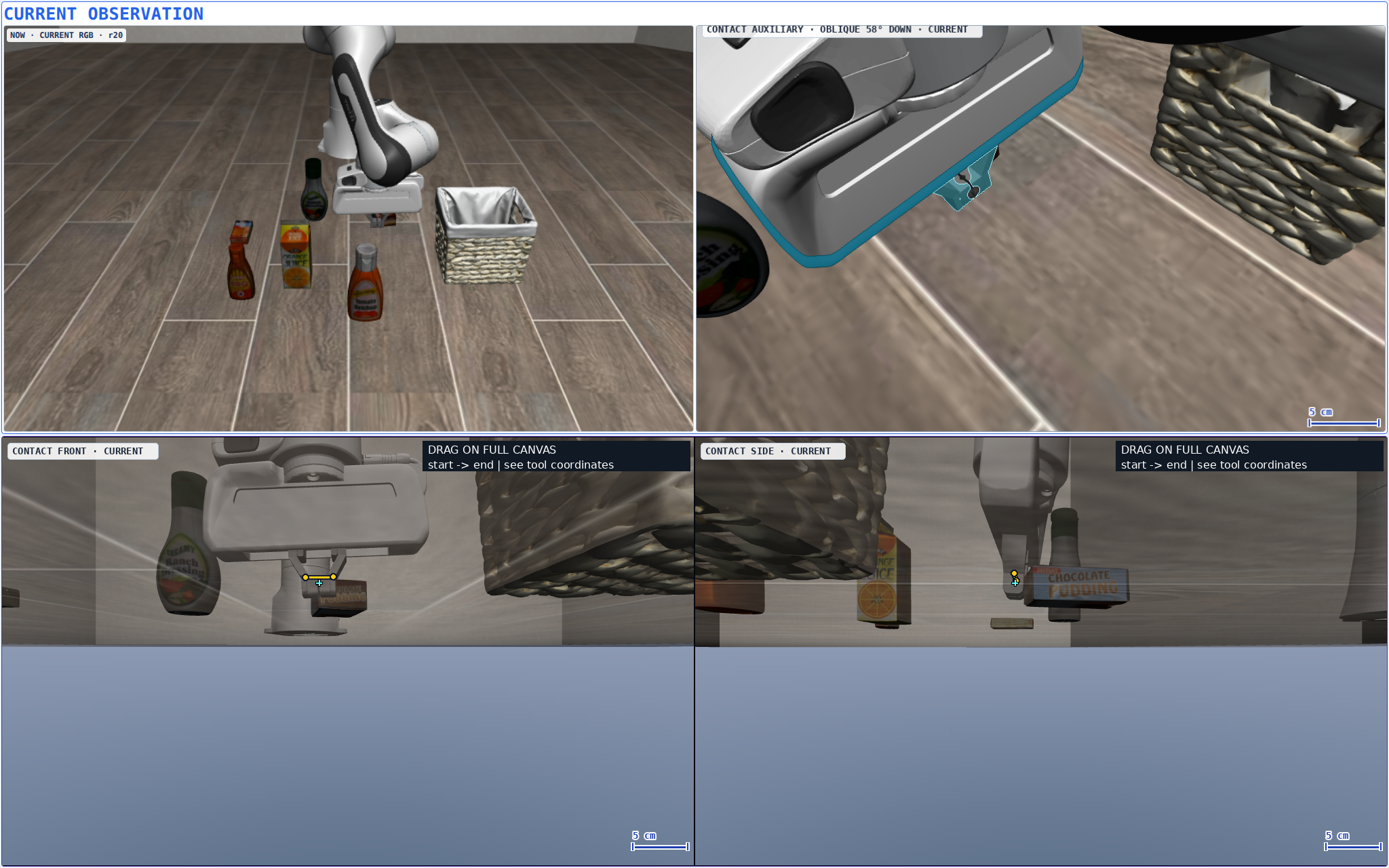}\\
        \footnotesize Turn 33: empty closure.
    \end{minipage}

    \medskip
    \begin{minipage}{\linewidth}
    \small\textbf{(b) Object identity tracking.}
    The task specifies the patterned bowl initially on the cabinet. After losing track of it, the agent redirected its approach to the bowl on the stove and later attempted to grasp the ramekin. It did not consistently preserve the instructed object's identity across scene changes and repeated localization.
    \end{minipage}\par\smallskip
    \begin{minipage}[t]{0.32\linewidth}\centering
        \includegraphics[width=\linewidth,viewport=0 640 1024 1244,clip]{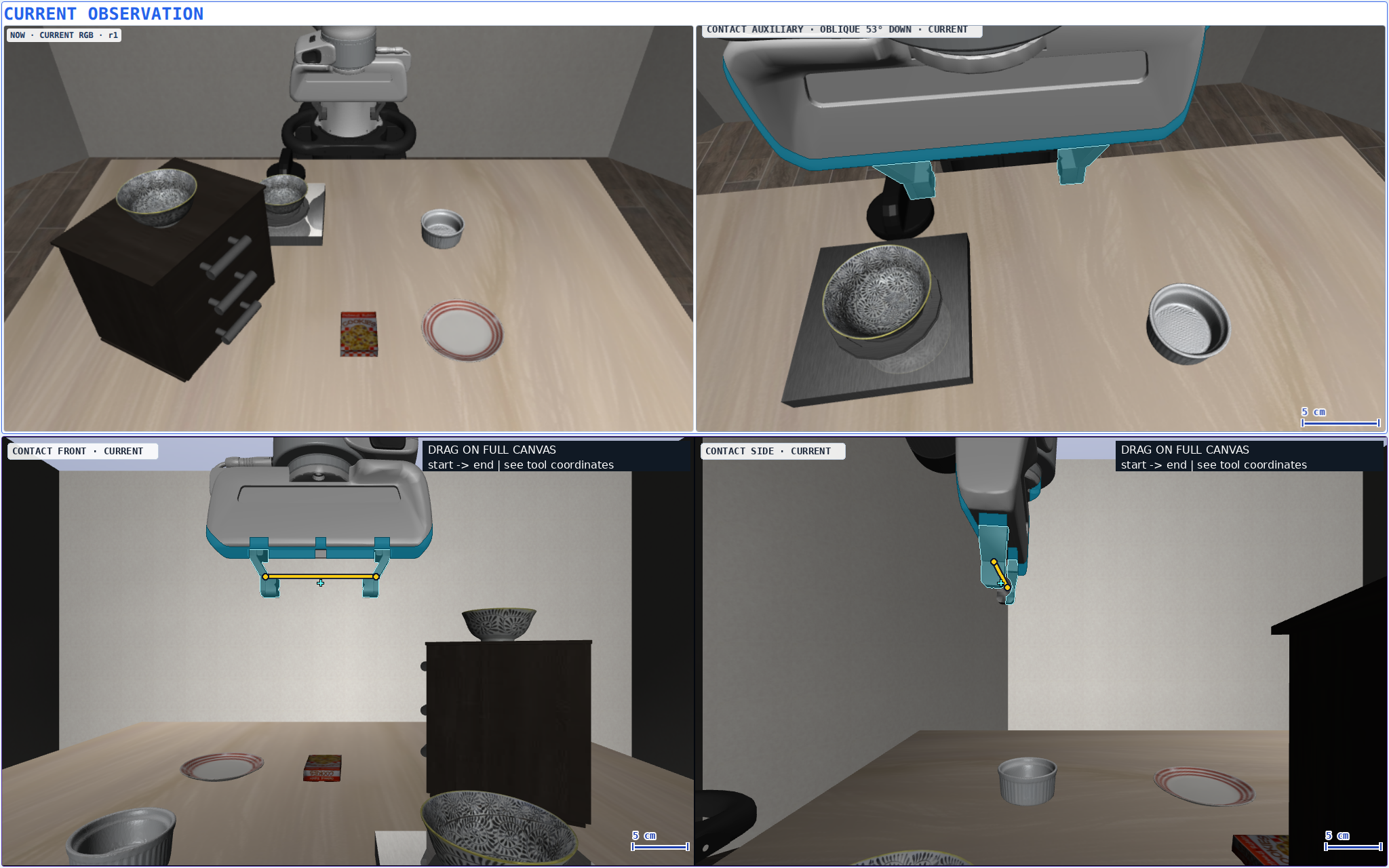}\\
        \footnotesize Turn 1: target on cabinet.
    \end{minipage}\hfill
    \begin{minipage}[t]{0.32\linewidth}\centering
        \includegraphics[width=\linewidth,viewport=1024 640 2048 1244,clip]{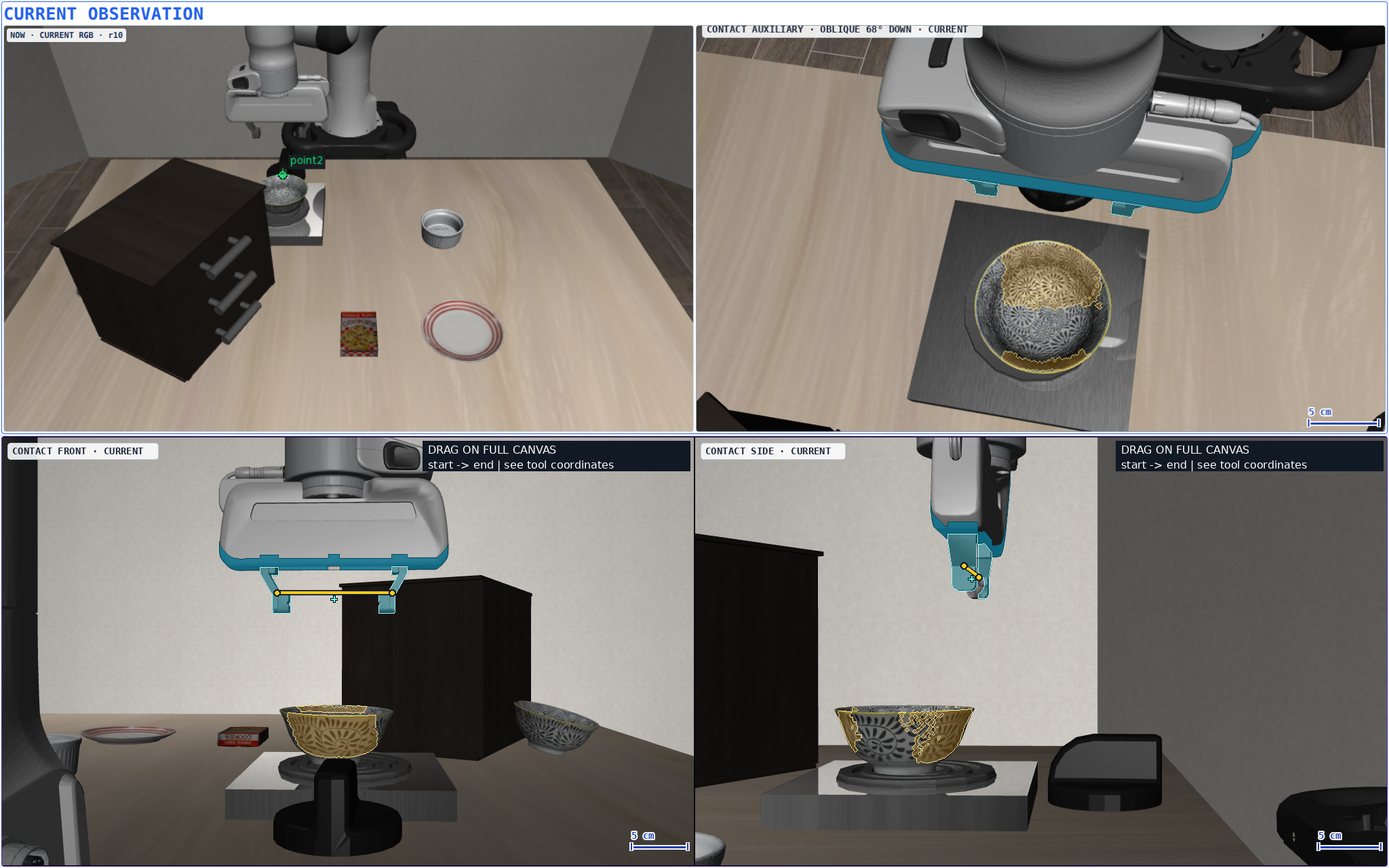}\\
        \footnotesize Turn 22: another bowl.
    \end{minipage}\hfill
    \begin{minipage}[t]{0.32\linewidth}\centering
        \includegraphics[width=\linewidth,viewport=1024 640 2048 1244,clip]{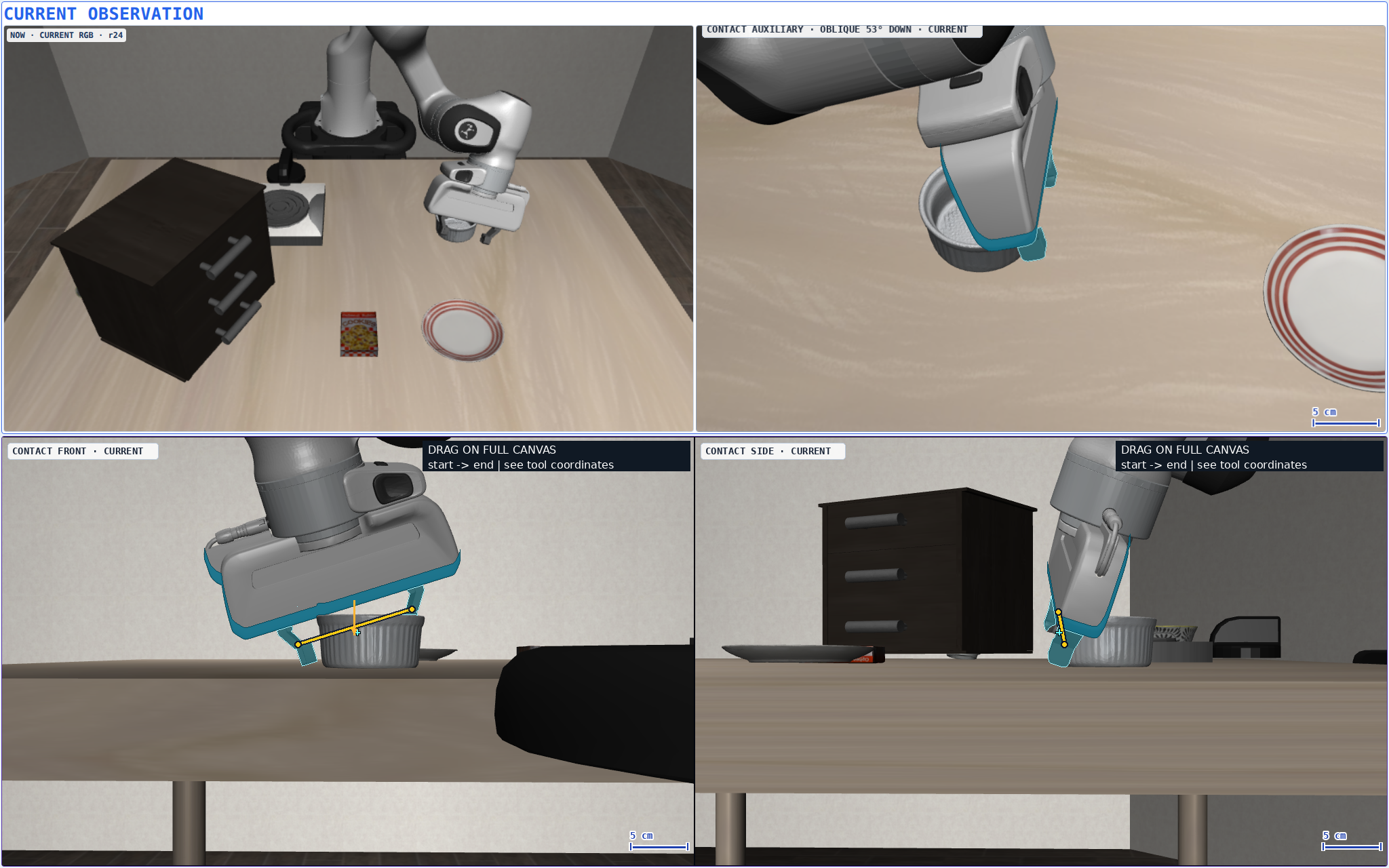}\\
        \footnotesize Turn 41: the ramekin.
    \end{minipage}

    \medskip
    \begin{minipage}{\linewidth}
    \small\textbf{(c) Premature completion judgment.}
    The agent released the bowl after describing it as centered and supported. After release, the bowl remained offset toward the plate edge, yet the agent continued to describe the placement as centered and complete. The task remained unsuccessful. This case illustrates insufficient geometric verification of the final object--target relation.
    \end{minipage}\par\smallskip
    \begin{minipage}[t]{0.32\linewidth}\centering
        \includegraphics[width=\linewidth,viewport=1024 640 2048 1244,clip]{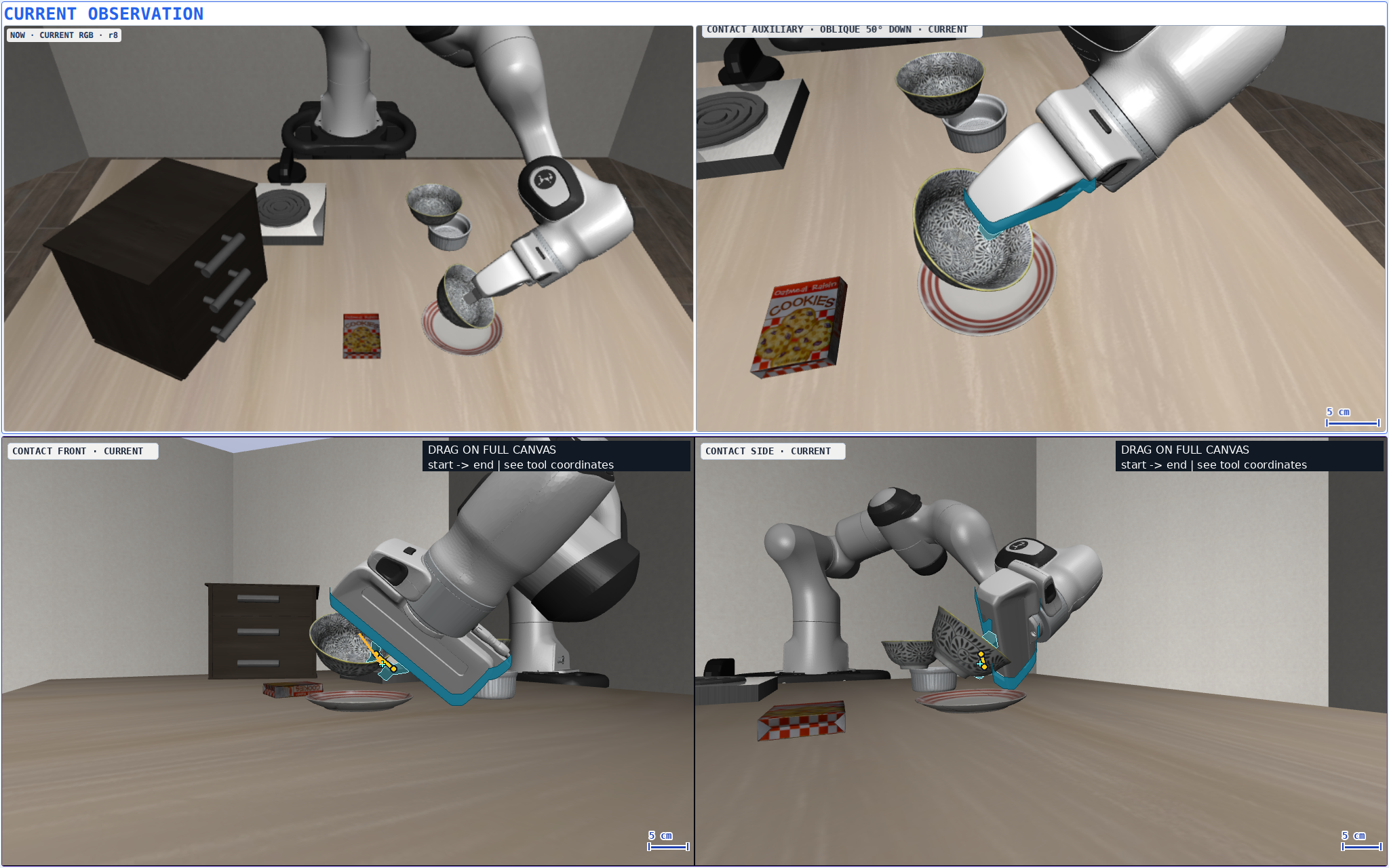}\\
        \footnotesize Turn 16: before release.
    \end{minipage}\hfill
    \begin{minipage}[t]{0.32\linewidth}\centering
        \includegraphics[width=\linewidth,viewport=1024 640 2048 1244,clip]{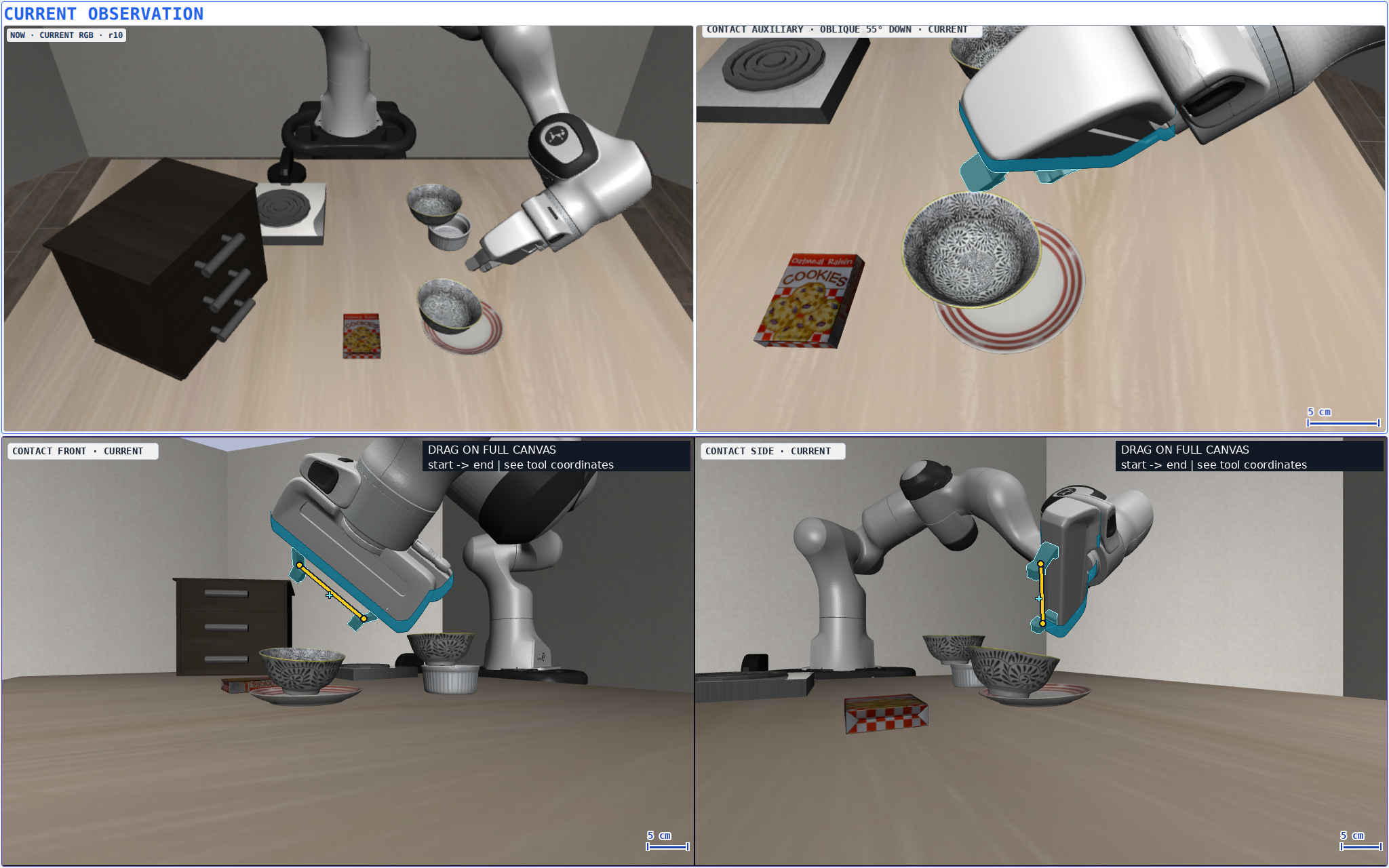}\\
        \footnotesize Turn 18: residual offset.
    \end{minipage}\hfill
    \begin{minipage}[t]{0.32\linewidth}\centering
        \includegraphics[width=\linewidth,viewport=1024 0 2048 640,clip]{failure_cases/placement_after.png}\\
        \footnotesize Turn 18: Side view.
    \end{minipage}
    \caption{Three failure cases. Panels are cropped from the Canvas received at the labeled decision turns.}
    \label{fig:failure-cases}
\end{figure}

We inspect three unsuccessful episodes from the 200-episode Object/Spatial evaluation. The observations below highlight difficulties in cross-view spatial judgment, object identity tracking, and verification of task completion. Figure~\ref{fig:failure-cases} shows the corresponding observations available to the agent.

\end{document}